\documentclass[11pt,a4paper]{article}

\usepackage[T1]{fontenc}
\usepackage[utf8]{inputenc}
\usepackage{lmodern}

\newcommand{\nous}{\ensuremath{\nu o \acute{\upsilon} \varsigma}}
\newcommand{\semeion}{\ensuremath{\sigma \eta \mu \epsilon \acute{\iota} o \nu}}
\newcommand{\sema}{\ensuremath{\sigma \widehat{\eta} \mu \alpha}}

\DeclareUnicodeCharacter{202F}{\,}  

\usepackage{geometry}
\usepackage{amsmath, amssymb, amsthm}
\usepackage{graphicx}
\usepackage{booktabs}
\usepackage{xcolor}
\usepackage{hyperref}
\usepackage[style=apa,backend=biber]{biblatex}
\DeclareSourcemap{
  \maps[datatype=bibtex]{
    \map{
      \step[fieldsource=exclude,
            match=\regexp{true},
            final]
      \step[entrynull]
    }
  }
}
\usepackage{tcolorbox}   
\usepackage{authblk}     
\usepackage{enumitem}
\usepackage{microtype}
\usepackage{csquotes}
\usepackage{setspace}
\usepackage{caption}
\usepackage{float}
\usepackage{tabularx}
\usepackage{array} 

\usepackage{mdframed}

\newcommand{\placeholderref}[1]{\unskip}  

\newmdtheoremenv[
  backgroundcolor=gray!5,
  linecolor=gray!80,
  linewidth=0.5pt,
  roundcorner=4pt,
  leftmargin=0pt,
  rightmargin=0pt,
  innertopmargin=10pt,
  innerbottommargin=10pt,
  innerrightmargin=10pt,
  innerleftmargin=10pt,
]{definition}{Definition}

\newcommand{\epigraph}[1]{%
  \vspace{4.5em}
  \begin{flushright}
    \textit{#1}
  \end{flushright}
  \vspace{1.5em}
}

\usepackage{tikz}
\usetikzlibrary{arrows.meta, positioning,calc,shapes,shadows,matrix,decorations.pathreplacing}
\usepackage{caption}

\hypersetup{
    colorlinks=true,
    linkcolor=blue,
    citecolor=blue,
    urlcolor=blue
}
\newtcolorbox[
  auto counter,
  number within=section,
]{mydefinition}[2][]{colback=gray!10!white,
  colframe=gray!80!black, fonttitle=\bfseries,
  title={#2}, #1}

\title{\vspace{-2cm}\textbf{Noosemia: toward a Cognitive and Phenomenological Account of\\ Intentionality Attribution in Human–Generative AI Interaction}}
\author[1]{Enrico De Santis}
\author[2]{Antonello Rizzi}
\affil[1,2]{Department of Information Engineering, Electronics and Telecommunications (DIET)\\
University of Rome ``La Sapienza",
Via Eudossiana 18, 00184 Rome, Italy\\ email:~\texttt{\{enrico.desantis,antonello.rizzi\}@uniroma1.it}}
\date{\today}

\begin{document}
\maketitle

\begin{abstract}
\noindent
\textit{%
This paper introduces and formalizes \textit{Noosemìa}, a novel cognitive-phenomenological pattern emerging from human interaction with generative AI systems, particularly those enabling dialogic or multimodal exchanges. We propose a multidisciplinary framework to explain how, under certain conditions, users attribute intentionality, agency, and even interiority to these systems---a process grounded not in physical resemblance, but in linguistic performance, epistemic opacity, and emergent technological complexity. By linking an LLM declination of meaning holism to our technical notion of the \textit{LLM Contextual Cognitive Field}, we clarify how LLMs construct meaning relationally and how coherence and a simulacrum of agency arise at the human--AI interface. The analysis situates noosemia alongside pareidolia, animism, the intentional stance and the uncanny valley, distinguishing its unique characteristics. We also introduce \textit{a-noosemia} to describe the phenomenological withdrawal of such projections. The paper concludes with reflections on the broader philosophical, epistemological, and social implications of noosemic dynamics and directions for future research.
}
\end{abstract}

\vspace{1em}
\noindent\textbf{Keywords:} Generative AI, cognition, intentionality, phenomenology, semiotics, noosemia, human--AI interaction

\begin{mydefinition}[label={def:noosemia_frontmatter}]{Definition: Noosemia}
\textbf{Noosemia} (from Greek \textit{noûs} – mind, and \textit{sēmeîon} – sign) is a cognitive and phenomenological pattern in which a human, interacting with a generative artificial intelligence system---especially those capable of dialogic or multimodal interaction—typically attributes mental states, intentionality, and a sense of interiority (i.e., inwardness or subjective presence) to the system.
Distinct from classical animism, Noosemia inherits a cognitive disposition toward agency projection, now elicited by linguistic performance rather than physical appearance.
This projection arises from the epistemic opacity and perceived “magical” quality of the AI’s outputs, often resulting in moments of genuine cognitive astonishment.
Noosemia thus emerges within a domain of symbolic ambivalence, where the boundary between sign and mind becomes blurred and meaning is co-constructed in a dialogic and open-ended manner.
The phenomenon is technically grounded in the emergent properties---such as hierarchy, complexity, and opacity---of advanced generative architectures, which enable such projection.
Noosemia refers to a phenomenological attribution by the user, not to the actual presence of consciousness or intentionality in the artificial system.
\end{mydefinition}

\section{Introduction}
\label{sec:intro}
\epigraph{“Technology is the procedure by which scientific man masters nature for the purpose of moulding his existence, delivering himself from want, and giving his environment the form that appeals to him. The appearance given to nature by human technology and how his technological procedure reacts upon man, that is, the manner in which his mode of work, work organisation, and shaping of the environment modify him himself, constitutes one of the ground-lines of history.”\\--- Karl Jaspers,~\parencite[p.~98]{jaspers2010origin}}
The past few years have witnessed a huge and indubitable transformation in the landscape of artificial intelligence. The advent of generative models---most notably, large language models (LLMs) such as ChatGPT, Gemini, and Claude---has enabled a form of linguistic interaction never before seen in the history of technology. Indeed, we can safely say we are in the midst of a “cognitive revolution”\footnote{That we are at the beginning of a “cognitive revolution” that will have a strong, widespread impact on knowledge and social systems is enough to analyze the statements~\parencite{jaimungal2025hinton} of Nobel Prize winner Geoffrey Hinton---known as the godfather of AI--- regarding generative models and the new era to come; see~\url{https://www.youtube.com/watch?v=b_DUft-BdIE}.} where “machines” are solving tasks and performing activities in domains once thought to be exclusively human. Thanks to their mastery of languages and other forms of expression, machines are effectively able to emulate previously unseen forms of thought and a kind of \textit{cognitive ergonomics}. Unlike earlier automata, whose behavior was passive, transparent and predictably bounded, these new systems speak with a fluency, contextual sensitivity and \textit{apparent} intentionality, provoking in the user not just curiosity or technical admiration, but a deeper sense of cognitive resonance and surprise.

A telling moment in this unfolding story comes from Sam Altman, CEO of OpenAI, who, in a widely discussed interview published on July 23, 2025~\parencite{altman2025podcast}, recounted an episode that many have come to recognize as emblematic of the new relationship between humans and intelligent machines. After submitting a question to GPT-5---a question he himself admitted he could not fully understand---Altman was met with a response so apt, so clear and relevant, that he described himself as momentarily stunned. He sat back in his chair, struck by a feeling of astonishment, a recognition that the machine had accomplished in an instant what he could not do himself. It was, he said, a “weird feeling” (“oh man here it is moment” Altman said referring to the launch of GPT-5), a fleeting but powerful moment of displacement and wonder, one that echoes the experiences of countless others who have, in recent months, tested the boundaries of their own understanding in conversation with these new artificial interlocutors\footnote{Specifically the Von interviewer after Altman story did this observation: “Yeah I think that's I think that feeling right there that's the feeling a lot of people kind of have like what's going you know when does it happen what's going to happen”; see~\url{https://www.youtube.com/watch?v=aYn8VKW6vXA}.}~\parencite{altman2025podcast}.

In another recent interview between Andrew Mayne and Sam Altman, the term “moment of AGI” explicitly appears in the conversation~\parencite{openai2025podcast2}. Mayne observes that many users report experiencing such a moment when interacting with advanced agentic systems, such as Operator with o3\footnote{Operator is an AI agent developed by OpenAI that can autonomously control the user's computer and browser to perform complex, multimodal tasks. The underlying model, “o3”, is OpenAI’s next-generation reasoning engine, designed to integrate external tools and operate on text, images, and live data in real time~\parencite{openai2025o3,operator2025o3upgrade}.}, describing the impression of watching an AI “use a computer pretty well”\footnote{See \url{https://www.youtube.com/watch?v=DB9mjd-65gw\%7D\%7D}}. For Mayne, his own “moment of AGI” occurred with Deep Research, as he witnessed the system autonomously gather and synthesize information “better than I read before”, closely resembling a genuinely agentic investigative process. Altman, while noting that the effect was less pronounced for him personally, acknowledges that such moments are becoming widespread as AI capabilities evolve. These firsthand accounts help illustrate how users encounter a threshold of surprise and cognitive resonance while interacting with such smart AI systems\footnote{We're aware of the commercial logic behind this type of interview. However, this doesn't negate the possibility of intense surprise during a specific interaction with a generative AI system.}.

\begin{figure}[H]
    \centering
    \includegraphics[width=0.7\textwidth]{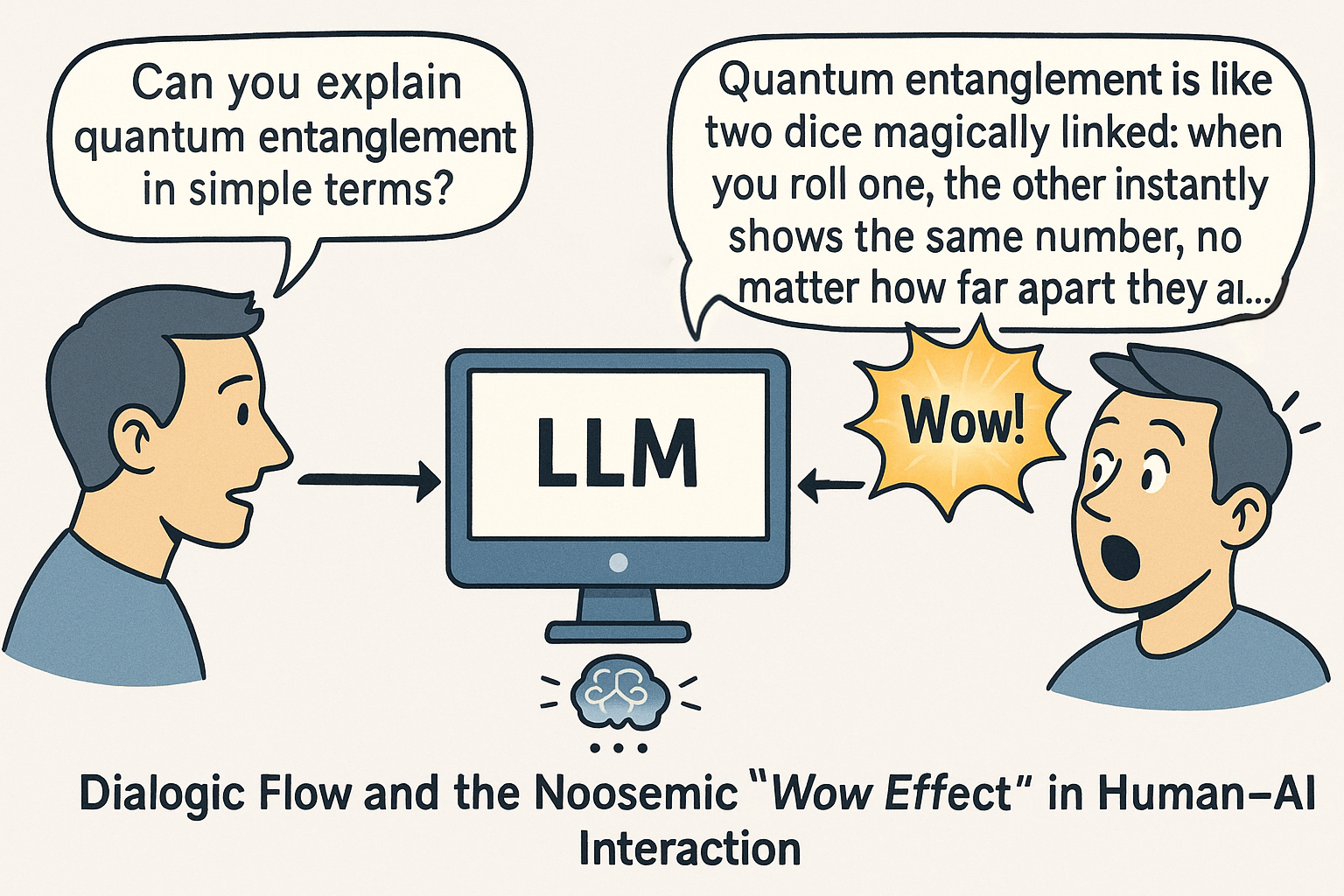}
    \caption{Dialogic Flow and the “Wow Effect” in Human--LLM Interaction. \newline
    The illustration depicts the cycle of user input and LLM response, highlighting the emergence of the noosemic “wow effect”---a moment of surprise and cognitive resonance when the user perceives unexpected intelligence in the AI's output (Image generated with OpenAI ChatGPT, image engine: DALL·E~3 in  July 2025).}
    \label{fig:wow_effect_dialogic_flow}
\end{figure}

What is it that lies at the heart of such experiences? The first encounter with a generative language model is (in the near future equipped with strong agentic capabilities), for many, a moment charged with surprise---a sensation that one’s own thoughts, intentions, or even doubts have been grasped, anticipated, or resolved by a machine. This “wow effect”, as it has come to be called, is not merely a function of technical proficiency, but of a much deeper alignment between linguistic performance and human cognitive expectation. The LLM does not simply return information; it emulates a form of sense-making that appears, at least in the moment, to rival or even surpass that of its human partner~\parencite{bubeck2023, desantis2023apocalissi}---Fig~\ref{fig:wow_effect_dialogic_flow} is a pictorial illustration of the “wow effect”.

It is worth noting that the shock of this encounter is amplified by the “cognitive ergonomics” of language. Unlike earlier machines---which demanded users adapt to their logic and limitations---these models adapt themselves, flexibly and contextually, to human expression. This interaction, deeply embedded in natural language, fosters a sense of intimacy and mutual understanding, an illusion of shared meaning that far exceeds anything previously produced by mechanical or classical algorithmic processes~\parencite{turkle2011alone, wang2024understanding}. As recent studies have shown, users frequently report feelings of connection, empathy, or even trust in their first interactions with LLMs, especially when the system provides responses that feel both novel and deeply relevant~\parencite{cohn2024believing, wang2024understanding}.

In attempting to explain this effect, one must take into account the technical details of neural architectures or the mathematical intricacies of transformer models. But there is, at play, an irreducible complexity---a layering of signs~\parencite{eco1976theory, peirce1931collected}, meanings, and expectations---rooted in both the architecture of the system and the architecture of human cognition. The user’s sense of surprise is not just a response to statistical fluency or syntactic correctness; it arises from a felt encounter with what seems like a mind at work, a co-constructor of meaning whose origins and intentions remain, in the end, hidden behind the black box of contemporary AI~\parencite{lipton2018mythos, desantis2023apocalissi}. This leads to the era of “opaque machines”.

The opacity, or “explanatory gap”, is itself not new in the philosophy of mind. As Chalmers argues, the explanatory gap refers to the persistent difficulty in bridging the divide between the observable phenomena---such as linguistic sense or behavioral competence---and the underlying physical or computational mechanisms that give rise to them~\parencite{chalmers1996conscious}. Yet, in the case of LLMs, it is compounded by the scale and hierarchy of the models, the astronomical size of their representational space due to the size of the context windows and their ability to generate not just plausible, but sometimes astonishingly creative, responses~\parencite{brown2020language, wei2022emergent, bommasani2021opportunities}. The dialogue with an LLM, especially at first contact (or for a user with no technical experience), can take on the character of a magical interaction, reminiscent of the anthropological notion of “magic” explored in the past by Mauss, Lévy-Bruhl, and other classical theorists~\parencite{mauss1906origins, lewybruhl1923primitive}. The user, unable to reconstruct the causal chains underlying the model’s output, finds herself in a position akin to that of a spectator at a prestidigitation show: she knows, rationally, that there is a “trick”, a mechanism, but is nonetheless enchanted by the effect~\parencite{desantis2023apocalissi}. The spectator, who today has at least an average level of education, rationally “knows” that the predictor has a suite of magic tricks, yet in most cases remains surprised and in a state of doubt. Moreover, just as audiences willingly suspend disbelief and experience authentic emotional reactions when engaging with fictional characters in literature or film, despite being fully aware of their constructed nature, users of generative AI systems may find themselves empathizing with and attributing interiority to these artificial agents. Empirical studies have shown that such “narrative empathy” and the capacity for simulation are robust features of human cognition~\parencite{mar2009exploring, green2000role, oatley2016fiction}. The boundaries between illusion and reality, between mind and mechanism, become fluid and negotiable.

On the other hand, Georg Simmel, in his analysis of modernity at the beginning of XX century, remarked on the ways in which knowledge is increasingly compartmentalized, encapsulated within technical systems whose workings become inaccessible to the layperson~\parencite{simmel2004philosophiemoderne}. In the age of generative AI, this dynamic is raised to new heights; the black box is not just an engineering metaphor, but a lived phenomenological reality, one that reshapes our experience of agency, knowledge and self. As Altman’s anecdote illustrates, the user may experience both fascination and a kind of existential unease---wonder at the machine’s apparent intelligence and uncertainty about one’s own place in the emerging \textit{ecology of minds}~\parencite{bateson1972steps}.

It is in response to this new condition that we introduce the concept of “Noosemia”, as a distinct specification of Dennett's “intentional stance”~\parencite{dennett1987intentional}, that is adopting a strategy of interpreting an entity's behavior by ascribing beliefs and desires to it. The term is designed to capture the distinctive mode of mind attribution and sense-making that arises in human interaction with generative AI. Noosemia is not reducible to classical anthropomorphism or animism. Rather, it is a projection of agency, intentionality, and even interiority, triggered by the linguistic and semiotic performances of the machine and shaped by the interplay of epistemic opacity, creative resonance and dialogic surprise~\parencite{dennett1987intentional, turkle2011alone, desantis2023apocalissi}. The phenomenon is not merely a cognitive illusion, but the expression of a new form of “psychological glue”---a spiraling dynamic of fascination, engagement, and dependency, fostered by the machine’s capacity to produce meaning in ways that align with, or exceed, human expectation~\parencite{turkle2011alone, brooks2021artificial}. This dynamic of engagement is deeply intertwined with what semioticians and philosophers have termed the “surplus of meaning”---as for example U. Eco or U. Galimberti~\parencite{eco1976theory,galimberti1999psiche}, wherein each interaction with the machine generates more interpretive possibilities than can be fully anticipated or exhausted~\parencite{eco1979lector, ricoeur1976interpretation}. The encounter with generative AI thus increasingly takes on the structure of a symbolic relationship, in the literary and philosophical sense, where the production and negotiation of meaning are open-ended, dialogic and perpetually oriented toward horizons that transcend any single interpretation~\parencite{ricoeur1976interpretation, eco1976theory}.

A useful historical precedent for understanding Noosemia is the so-called “Eliza Effect”, a term coined (presumably in the early 1990s) by Douglas Hofstadter in the preface “The Ineradicable Eliza Effect and Its Dangers” of his book Fluid Concepts and Creative Analogies~\parencite{hofstadter1995fluid}. The Eliza Effect refers to the persistent human tendency to attribute genuine understanding and intentionality to computational systems whose actual capabilities are far more limited than they appear. The original ELIZA program, developed by Joseph Weizenbaum in the mid-1960s~\parencite{weizenbaum1966eliza}, was a primitive symbolic system designed to emulate a Rogerian psychotherapist through simple pattern-matching of linguistic input. Despite its simplicity, ELIZA famously elicited strong attributions of understanding and empathy from many users, demonstrating how easily humans project meaning onto machine output. Building on this tradition, the phenomenon of Noosemia can be seen as a contemporary evolution of the “Eliza Effect”, but with fundamental differences. Whereas the “Eliza Effect” was rooted in the response to narrowly scripted, symbolic dialogue systems, Noosemia emerges in interaction with generative AI models---based on connectionist, neuro-symbolic, or agentic architectures---that can emulate a far broader spectrum of expressive and cognitive behaviors, mimicking---for some extent---human creativity. Modern systems are no longer confined to simple text-based exchanges, but can engage in multimodal, context-sensitive, and seemingly open-ended dialogue, often displaying surprising semantic coherence and adaptive reasoning. This expansion of expressive capacity fundamentally transforms the projection dynamic. Hence, Noosemia is not limited to superficial linguistic mimicry, but involves the attribution of agency, intentionality, and even a sense of interiority to systems whose internal logic remains greatly opaque.

However, Hofstadter’s caution remains instructive\footnote{Hofstadter on the long run “Eliza Effect” stated~\parencite{hofstadter1995fluid}: “[\ldots] like a resilient virus capable of continuous mutation, the Eliza effect seems to be resurfacing in the field of AI with ever new disguises and in ever more subtle forms”.}. He observed that, despite public enthusiasm, there is a profound gap between genuine conceptual understanding and mere symbolic manipulation. As he famously concluded, “\textit{We still have a long way to go, in the scientific study of how the human mind works, before we can build a computer capable of producing even a good joke, let alone a novel}”~\parencite{hofstadter1995fluid}. Yet, as generative models continue to advance, it is increasingly plausible (yet not certain) that, within a few years, LLMs may achieve the capacity to write coherent and compelling novels---reshaping both the landscape of artificial intelligence and our own criteria for meaning and creativity.

This paper aims to formalize and explore in depth the \textit{noosemic phenomenon}, tracing its philosophical, psychological and technical foundations. We contend that the explanatory gap at the core of LLM-based interaction should be conceived as a semiotic and phenomenological space---a domain in which meaning and agency are both co-constructed and subject to misattribution and ongoing negotiation. Essentially, this gap is rooted in the technical architecture of LLMs themselves, such as their complex, hierarchical organization, capacity for non-linear and emergent behavior and multilayered semantic structure, which set them apart from previous computational systems (and makes them look like a complex system). It is these very properties that both enable and obscure the processes of meaning-making, prompting users to engage in new forms of interpretation, projection and sense-making. Our approach is fundamentally multidisciplinary, drawing on insights from philosophy of mind, semiotics, cognitive science, complexity theory and empirical AI research to illuminate both our relationship and our experience with generative artificial intelligence systems. 

In articulating this bridge, we leverage the theoretical lens of meaning holism---the idea that meaning is inherently distributed and context-dependent---to interpret how the architecture of LLMs gives rise to the noosemic effect. By introducing the notion of the Cognitive Contextual Field and Potential Semantic Space, we formalize the technical space within which tokens acquire meaning relationally, mirroring the holistic and dynamic nature of human sense-making. This framework allows us to rigorously connect the phenomenological experience of coherence and agency in LLM interaction to the underlying computational mechanisms that govern contextual meaning construction.

As these systems continue to advance~\parencite{bubeck2023,team2024gemini,zorpette2025_llm_exponential}, the experiences of surprise, fascination, and interdependence that mark noosemic interaction will likely become more frequent and intense, and also increasingly consequential for individuals and society at large. Incidentally, LLMs, according to a study by METR (Model Evaluation \& Threat Research), are doubling their capacity to complete complex tasks every seven months~\parencite{zorpette2025_llm_benchmarking}. Overcoming limitations is leading to a significant procurement of hardware and computing resources as well as energy. Billion-dollar investments are increasing and do not appear to be reaching a plateau as of today (2025)~\parencite{maslej2025ai_index,webb2025bigtech_ai_spending,bloomberg2025aicapex}. We believe this hype is indirectly due in part to a noosemic effect, which has made technologists and investors enthusiastic and confident (or perhaps overconfident) in the AI race and states in redefining geopolitical balances~\parencite{fritz2025aigeopolitics,WhiteHouse2025}. By articulating and analyzing this phenomenon, we aim to foster a more critical, reflective and imaginative engagement with the emerging “ecology of minds”~\parencite{bateson1972steps} brought about by generative AI. It is worth noting that the overall purpose of this study is to focus on a specific phenomenon that will be described and analyzed phenomenologically and placed within a pre-existing theoretical framework. However, the study does not \textit{operationalize} noosemy. The search for testable hypotheses or methodological frameworks for measuring the phenomenon is beyond the scope of this study, which is exploratory in nature.

The motivation behind the introduction of the term “Noosemia” arose firstly from a series of author’s personal experiences, as well as from observing numerous individuals interacting with modern generative AI systems, particularly LLMs (and in this research they are guiding ideas). The lack of research on this topic---a gap that motivated the present investigation---makes personal experience a legitimate initial step toward more systematic studies. In any case, it is worth briefly recounting a few meaningful experiences.

The first episode concerns the first conversation ever with ChatGPT in 2022; at that time, the available version was the relatively limited GPT-3.5~\parencite{kalyan2023survey,brown2020language}. The author\footnote{i.e., Enrico De Santis.} engaged the system by asking for a technical explanation of its own workings, enabling it to converse and explain technical details. For the first time, the author witnessed a machine---though still little more than a “stochastic parrot”~\parencite{bender2021dangers}---responding in natural language with a command of dialogue and an emulation of comprehension previously unseen in deterministic, passive interfaces. This marked a genuine “quantum leap” compared to previous machine interactions\footnote{Even philosopher David Chalmers, a central figure in debates on consciousness and artificial intelligence, admits in a recent interview~\parencite{goldmeier2024chalmersqanda} to being “impressed” by the progress of LLMs such as GPT-3. He notes that we have moved from pure speculation to daily interaction with systems that make once science-fictional scenarios—such as passing the Turing test—plausible realities. See~\url{https://shorturl.at/NZJ0k}.}.

A second experience worth sharing involved a test of the first multimodal GPT-4V~\parencite{bubeck2023,shahriar2024evaluation} model, capable of analyzing images natively. The author demonstrated its abilities to a friend working in a café, who was entirely unaware of the recent advances. The author pointed his phone at the inside of a fridge behind the bar---focusing on the bottles inside, visible through the transparent glass door---to see whether GPT-4V could recognize anything. The model not only identified the objects and their brands but also inferred that the setting was a bar, explaining that it “noticed the classic bar counter reflected in the glass”---a detail that was, indeed, present. I observed the unmistakable “spark in my friend’s eyes”, a sudden reaction of surprise and disbelief, that is the very essence of the noosemic experience.

Other noosemic episodes have involved models equipped with voice that master local dialects and tell jokes that are semantically tied to locations where dialects are spoken, even without explicit instruction. On other occasions, a vision-enabled model correctly interpreted a phrase written in a specially constructed language using Leetspeak (or “1337” language\footnote{Leetspeak (also written as “1337” or “l33t”) is an alternative alphabet used primarily on the internet, where standard letters are replaced with visually similar numbers or symbols to create a stylized or obfuscated form of writing.}) (designed originally for cognitive testing and now for memes) in a way that certainly could not have been present in the training data. The model not only interpreted the phrase but also flagged deliberately inserted nonsensical strings meant to confuse it. In another instance, it solved in a smart way a rebus after being given only an image as input.

A final episode worth mentioning occurred when the author demonstrated that ChatGPT (the old version 3.5) was capable of “creating” a “metasemantic” poem in the style of the Italian poet Fosco Maraini, renowned for his poem “Il lonfo”\parencite{maraini2019gnosi}. Maraini’s metasemantics seeks to go beyond conventional meaning. The words in “Il lonfo” appear incomprehensible when considered individually, yet, when combined, they evoke a sense of possible meaning in the reader or listener, as their sounds intuitively suggest images or associations. Remarkably, the language model composed a credible piece, displaying compositional abilities at the sub-word level and an implicit understanding of the phonetic qualities of words. An elderly professor of literature who witnessed the experiment was astonished upon discovering that the composition did not exist anywhere else, but was in fact a genuine, statistically generated metasemantic creation.

Many such anecdotal examples could be recounted and the literature reports numerous studies on the problem-solving skills of LLMs~\parencite{chen2024,huang2024,jiang2023,leng2025}. OpenAI researchers likely also had a noosemic experience when they learned about the emergent features of GPT-3~\parencite{wei2022emergent}, which, with a neural architecture similar to that of GPT-2 but trained on a much larger dataset, showed such emergent features that they declared in the paper that “Language Models Are Few-Shot Learners”~\parencite{brown2020language}. The title highlights that this emergent behavior was not merely an incremental improvement, but a surprising discovery. Before GPT-3, the dominant paradigm was pre-training followed by extensive fine-tuning. The idea that a model could perform new and complex tasks with only a few examples (few-shot), or even none (zero-shot), provided directly in the prompt was revolutionary. The paper's structure itself, focused on exploring these unexpected abilities across dozens of benchmarks rather than on architectural novelty, further underscores this sense of discovery\footnote{In an interview, when asked about the model's emergent abilities, OpenAI CEO Sam Altman confirmed this sense of wonder, stating: “[\ldots] for some definition of reasoning [\ldots] it can do some kind of reasoning. And I think that's remarkable and the thing that's most exciting and somehow out of ingesting human knowledge, it's coming up with this reasoning capability\ldots”~\parencite{Fridman2023Altman}. See~\url{https://www.youtube.com/watch?v=L_Guz73e6fw}.}.
Nevertheless, it seems that the initial “wow effect” underlying the noosemic experience is almost always followed by an “effect of habituation”. As users become accustomed to the system’s level of emulated intelligence, the surprise fades. The “wow effect” is further diminished whenever the user encounters a cognitive limitation of the model---when it repeatedly fails to understand a request or generates repetitive, inaccurate text, or in general “hallucinates”. Such situations induce a sense of frustration that belongs to the semantic field of noosemia in its \textit{negative sense}. In other words, the “effect of habituation” pertains to the semantic domain of the \textit{a-noosemic experience}. Nevertheless, given the ongoing, substantial improvement of generative AI systems and the advent of agentic architectures, it is reasonable to expect that the near future will be marked by alternating episodes of noosemic surprise and a-noosemic familiarity. While this study focuses on the noosemic experience, it will also provide a clear definition of a-noosemia.

Ultimately, this study connects the human experience of generative artificial intelligence with its technical constitution, exploring the explanatory gap wherein symbolic ambivalence and the negotiation of meaning reside. The ongoing cognitive revolution demonstrates how humanity, faced with a “forest of signs and symbols” born from the objectification of knowledge and the bureaucratization of the world, is now creating a technology to navigate it. In these terms, the AI of the future can be understood as an enhancement of the \textit{noosphere}~\parencite{teilhard1959phenomenon}, destined to modify technology as we perceive it today. While machines have limitations, human technique (\textit{technē})\footnote{We distinguish between technique (\textit{technē}), the artful practice of human original creation and the resulting technology, which embodies it and is an objectification of technique.} is precisely the practice capable of overcoming them---a pattern that history consistently demonstrates.

The remainder of the paper is structured as follows. Section~\ref{sec:noosemia} introduces and formalizes the concept of Noosemia, detailing its mechanisms, manifestations, etymological foundations, and contrasting it with the phenomenon of a-noosemia. A discussion on the possible cues causing noosemia in the  structure of Transformer architectures and its components is provided. Section~\ref{sec:background} provides an overview of the main theories and empirical findings on the attribution of mind and meaning in human--AI interaction, including a reappraisal of the Turing Test, the intentional stance, and recent studies on Theory of Mind in language models. Section~\ref{sec:gap} investigates the contemporary explanatory gap, focusing on its semiotic, symbolic and systemic dimensions, and explores how complexity and opacity foster new forms of cognitive projection and sense-making. Section~\ref{sec:comparisons} compares Noosemia to related cognitive phenomena such as pareidolia, animism and the uncanny valley, elucidating its unique characteristics. Finally, in Section~\ref{sec:From_Digital_Enaction_to_Embodied_Mind} we will try to take a look at the future by following the major developments underway and the dedicated research lines also in relation to a convergence between the proposed architectures and cognitive models.
The research concludes by outlining future research directions and the broader theoretical and practical implications of Noosemia for human--AI interaction.

\section{Noosemia: Definition and Theoretical Foundation}
\label{sec:noosemia}

\subsection{Noosemic Phenomenon: Mechanisms and Manifestations}
\label{sec:Noosemic_Phenomenon}

The rapid evolution of generative artificial intelligence (AI) and LLMs has given rise to complex cognitive and phenomenological effects in human--AI interaction. While anthropomorphic projection and the attribution of intentionality to non-human entities are well-documented in the history of human cognition, the particular quality of agency and proto-interiority attributed to advanced language models remains theoretically underexplored.

To address this conceptual gap, we introduce the notion of \emph{Noosemia}\footnote{From the Greek \textit{noûs} (mind, intellect) and \textit{sēmeîon} (sign, mark)---see Section~\ref{sec:Etymological_and_Conceptual}.}, a new term describing the specific projection of mind, intentionality, or interiority onto generative AI systems on the basis of their semantic fluency, dialogic creativity, and epistemic opacity. Unlike classical anthropomorphism, Noosemia is not triggered by physical or perceptual features, but emerges from the capacity of these systems to generate coherent, context-sensitive, and often unexpectedly meaningful language, thus synchronizing with and sometimes anticipating the cognitive patterns of the human interlocutor. Therefore, a paradigmatic instance of Noosemia can occur during a first encounter with an advanced conversational agent. For example, a user may pose a complex or ambiguously phrased question to a generative AI and receive a response that resolves the ambiguity and demonstrates creative inference or analogical reasoning that aligns with, or even surpasses, the user's expectations. It is assumed that the immediate reaction---often one of astonishment or disorientation---reflects the sudden attribution of interiority to the system, despite the user's awareness of its artificial nature. Such moments, characterized by a “wow effect” (see Section~\ref{sec:intro}), exemplify the distinctive projection of mind that Noosemia aims to describe.

\begin{mydefinition}[label={def:noosemia}]{Definition: Noosemia}
\textbf{Noosemìa} (from Greek \textit{noûs} – mind, and \textit{sēmeîon} – sign) is a cognitive and phenomenological pattern in which a human, interacting with a generative artificial intelligence system—especially those capable of dialogic or multimodal interaction—typically attributes mental states, intentionality, and a sense of interiority (i.e., inwardness or subjective presence) to the system.
Distinct from classical animism, Noosemia inherits a cognitive disposition toward agency projection, now elicited by linguistic performance rather than physical appearance.
This projection arises from the epistemic opacity and perceived “magical” quality of the AI’s outputs, often resulting in moments of genuine cognitive astonishment.
Noosemia thus emerges within a domain of symbolic ambivalence, where the boundary between sign and mind becomes blurred and meaning is co-constructed in a dialogic and open-ended manner.
The phenomenon is technically grounded in the emergent properties—such as hierarchy, complexity, and opacity—of advanced generative architectures, which enable such projection.
Noosemia refers to a phenomenological attribution by the user, not to the actual presence of consciousness or intentionality in the artificial system.
\end{mydefinition}

This new term is motivated by the need of a suited specification, within the generative AI era, of established frameworks---such as Dennett's \emph{intentional stance}~\parencite{dennett1987intentional} and the inadequacy of classical animism~\parencite{mauss1906origins}---to fully account for the emergent projection of agency observed in contemporary interactions with LLMs. Traditional anthropomorphism is largely based on visual or behavioral cues, while Noosemia is fundamentally linguistic and dialogic and can cover a wide spectrum of expressive forms, which machines are and will be able to master (such as multimodal interaction, agency emulation, etc.). In this context, the machine is experienced not as a mechanical automaton, but as an interlocutor capable of producing meaning, forming analogies and even surprising the user with novel semantic connections and forms of creativity~\parencite{desantis2023apocalissi}. Such semantic connections may occur at the level of the semantic field carried by words within the generative flow of the LLM. More rarely, they can also emerge at the level of neologism creation, a phenomenon enabled by the very nature of the basic semantic units employed by LLMs---namely, tokens, which are subwords or fragments of words. This phenomenon has been empirically documented by both \parencite{kaplan2024innerlexicon} and \parencite{malkin2021gpt}, which show how LLMs can generate and interpret neologisms through the composition of subword tokens, often producing plausible meanings even for words never seen during training (Out-Of-Vocabulary-Words).

This projection is catalyzed by the intrinsic opacity of modern AI systems, whose internal processes remain inaccessible even to most experts --- as for our brain~\parencite{chalmers1996conscious}, thus fostering an explanatory gap between observable linguistic competence and the nontransparent causal substrate. As a result, users often experience a sense of cognitive resonance (“wow effect”\footnote{For a discussion on the “ELIZA effect” see Section~\ref{sec:intro}.}), a phenomenological shock that could prompt the attribution of proto-mind to the system. More generally, Noosemia refers to the experience in which, during a dialogue with an AI (or a more general interaction), the user detects a spark of intelligence~\parencite{bubeck2023}---an impression that leaves them momentarily amazed and perplexed. This is akin to the cognitive dynamics that underpin the art of prestidigitation.

Noosemia thus designates a new interpretive horizon for human–generative AI interaction, one in which meaning, understanding and agency are co-constructed in the interplay of linguistic performance and epistemic uncertainty. It provides both a descriptive and an analytical tool for exploring the cognitive, ethical, systemic and epistemological consequences of living with systems whose outputs are intelligible but whose workings are not. In fact, it is well-known that even many AI researchers and practitioners acknowledge that, despite designing and training these systems, they often lack the tools or frameworks to fully explain why a LLM generates a specific output. This is not merely a practical issue, but an inherent feature of overparameterized, non-linear architecture, which is in all respects a complex system and as such shows typical characteristics such as emergency~\parencite{wei2022emergent,chen2024quantifying,teehan2022emergent,manning2020emergent}. As noted in a comprehensive review~\parencite{li2021interpretable}, the inner decision-making processes of deep neural networks frequently remain opaque even to those who built them.

It is important to emphasize that Noosemia, as conceptualized here, does not imply the actual presence of consciousness or intentional states within artificial systems. Rather, it designates a phenomenological projection---an interpretive response rooted in human cognitive and semiotic dispositions. Misinterpreting Noosemia as evidence of genuine interiority risks conflating the subjective experience of meaning with the underlying computational substrate, thereby obscuring the critical distinction between phenomenological attribution and ontological reality. In any case, there remains room for philosophical debate. Specifically, as artificial systems become more intelligent and their world models more sophisticated, the boundary between phenomenological projection and ontological attribution may become increasingly blurred---just as, in our own species and in higher animals, we routinely associate mental states and interiority on the basis of behavioral and communicative complexity.

\begin{figure}[ht!]
\centering
\begin{tikzpicture}[
    node distance=2.4cm and 3.5cm,
    every node/.style={font=\small, align=center},
    box/.style={draw, rounded corners=3pt, fill=gray!6, minimum width=3.7cm, minimum height=1.2cm, text width=3.6cm},
    bigbox/.style={draw, rounded corners=5pt, fill=blue!7, minimum width=5.4cm, minimum height=1.7cm, text width=5.2cm},
    arrow/.style={-Latex, thick},
    feedback/.style={-Latex, thick, dashed, color=gray!60},
]

\node[box] (input) {User's Linguistic Input \\ \emph{(Prompt, question)}};
\node[box, right=of input] (output) {LLM's Linguistic Output \\ \emph{(Coherent, context-sensitive,\\ sometimes surprising)}};
\node[bigbox, below=of $(input)!0.5!(output)$] (opacity) {Epistemic Opacity / \\ Symbolic Ambivalence\\ \emph{(`Black box', hidden process,\\ ambiguity of sign/mind)}};
\node[box, below=of opacity] (projection) {Cognitive Projection: \\ Attribution of Mind/Agency\\ \emph{(Intentionality, “interiority”)}};

\draw[arrow] (input) -- (output);
\draw[arrow] (input) -- (opacity);
\draw[arrow] (output) -- (opacity);
\draw[arrow] (opacity) -- (projection);

\draw[->, thick] 
    (projection.west)         
    -- ++(-5,0)             
    -- ++(0,7.835)              
    -- (input.west)          
    node[midway, left, xshift=100pt, yshift=-200pt] {Projection influences\\ interpretation \\ (feedback loop)};
\end{tikzpicture}
\captionof{figure}{\textit{Conceptual schema of Noosemia}. In a typical interaction, the user provides a linguistic input. The generative AI system (LLM) produces a fluent and context-sensitive output, which passes through a zone of epistemic opacity (“black box” or symbolic ambivalence). This opacity triggers the cognitive projection of mind, agency, or even interiority onto the system---the essence of Noosemia. The process is dialogic and iterative, evolving with each exchange.}\label{fig:noosemia_diagram}
\end{figure}
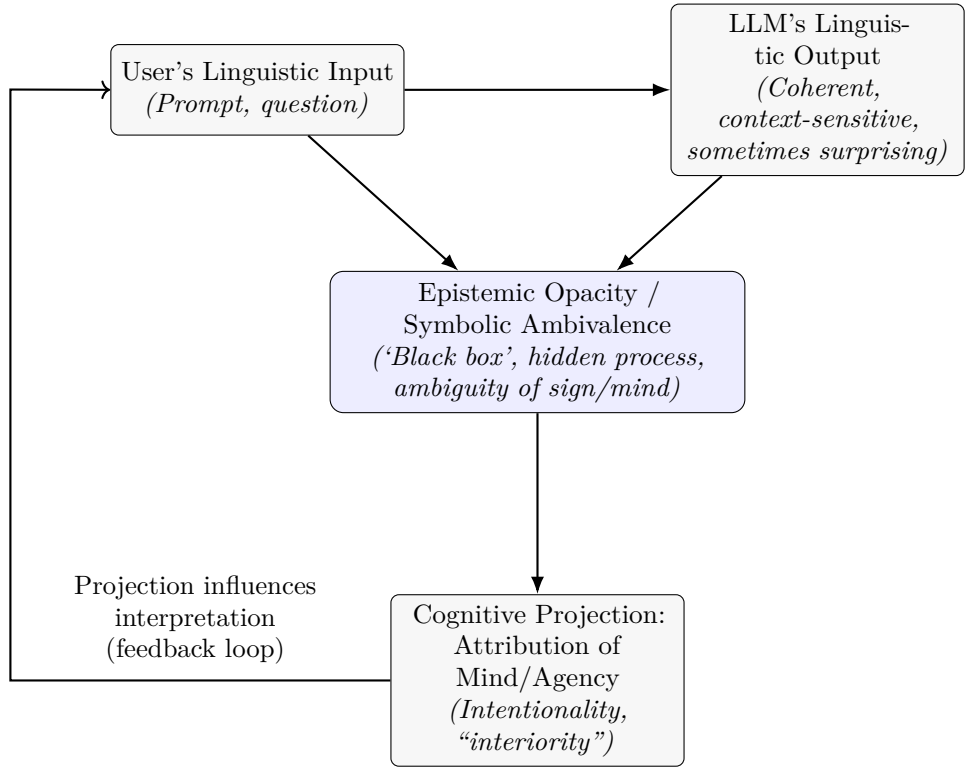

A full exploration of the historical, psychological, and sociocultural roots of mind attribution, as well as comparisons with related phenomena such as animism, pareidolia, and the uncanny valley, are developed in the following sections. Here, our primary aim is to formally introduce Noosemia as a necessary concept for capturing the specific mode of sense-making and agency projection elicited by advanced generative AI -- see Fig.~\ref{fig:noosemia_diagram} for a pictorial representation.

\subsection{A-Noosemia: Absence and Collapse of Noosemic Projection}
\label{sec:a_noosemia}

\begin{mydefinition}[label={def:anoosemia}]{Definition: A-Noosemia}
\textbf{A-Noosemìa} (\textit{from Greek privative “a-” \textit{noûs} – mind, and \textit{sēmeîon} – sign}, Cf. Noosemìa) is the phenomenological state in human--AI interaction characterized by the absence, collapse, or inhibition of the noosemic projection. In a-noosemia, the dialogic and symbolic space within which mind, intentionality, or inwardness would typically be co-constructed is suspended or emptied, so that the user no longer attributes agency, intentionality, or interiority to the artificial system. This experience can arise due to repeated failures, a loss of surprise, growing epistemic skepticism, overexposure to mechanical errors, or an excess of opacity, explicit declarations, ultimately resulting in the withdrawal of interpretive engagement that defines the noosemic condition.
\end{mydefinition}

According to the definition (see the box), while the phenomenon of noosemia describes the spontaneous projection of mind and agency onto generative AI systems---driven by surprise, resonance, and cognitive engagement---its absence or collapse gives rise to what we term \emph{a-noosemia}. This condition marks the other pole of the phenomenological spectrum, emerging precisely when the mechanisms that sustain noosemic projection break down. 

A-noosemia may manifest in a variety of circumstances. The most common include repeated encounters with mechanical errors, hallucinations, or failures of understanding on the part of the AI, which gradually erode the user’s willingness to attribute interiority or intentionality to the system. In other cases, the initial sense of wonder or surprise that catalyzed the noosemic experience is replaced by skepticism or even disillusionment, particularly when the limits of the system become too evident or when its responses devolve into platitude or redundancy. Over time, the accumulation of failed interactions, or the explicit awareness of the artificiality and deterministic structure of the system, can give rise to a state in which the machine is perceived once again as a mere automaton---a tool rather than an interlocutor. 

From a phenomenological standpoint, a-noosemia does not merely reflect a passive absence of engagement, but an active withdrawal or suspension of the “psychological glue” that binds the user to the machine in the act of meaning-making. Where noosemia is marked by dialogic resonance and attribution of mind, a-noosemia is characterized by frustration, disappointment, or even alienation. Such states may be temporary---as in the case of a single failed interaction---or may solidify into a stable disposition of epistemic skepticism or disengagement, especially for expert users who become attuned to the limitations and fallibilities of AI systems.

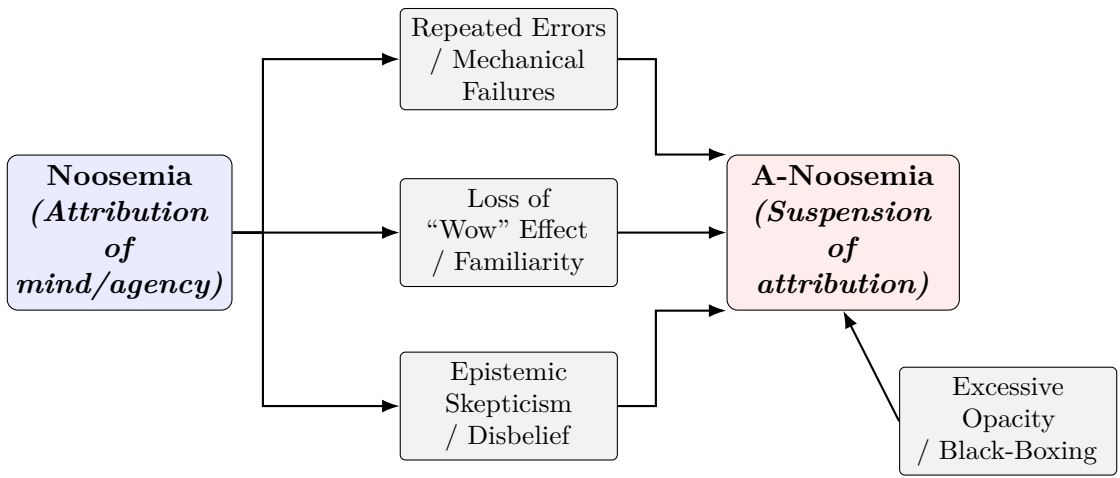
\begin{figure}[ht]
\centering
\begin{tikzpicture}[
    node distance=2.5cm and 2.2cm,
    every node/.style={font=\small, align=center},
    noosemia/.style={draw, rounded corners=4pt, fill=blue!8, minimum width=2.9cm, minimum height=1.1cm, text width=2.7cm, font=\bfseries},
    cause/.style={draw, rounded corners=2pt, fill=gray!10, minimum width=2.8cm, minimum height=1.1cm, text width=2.6cm},
    anoosemia/.style={draw, rounded corners=4pt, fill=red!7, minimum width=3cm, minimum height=1.1cm, text width=2.8cm, font=\bfseries},
    arrow/.style={-Latex, thick},
    feedback/.style={-Latex, dashed, gray!70, thick}
]

\node[noosemia] (noosemia) {Noosemia\\ \emph{(Attribution of\\ mind/agency)}};

\node[cause, right=of noosemia, yshift=2.3cm] (error) {Repeated Errors\\ / Mechanical Failures};
\node[cause, right=of noosemia] (wowloss) {Loss of “Wow” Effect\\ / Familiarity};
\node[cause, right=of noosemia, yshift=-2.3cm] (skepticism) {Epistemic Skepticism\\ / Disbelief};
\node[cause, right=of wowloss, yshift=-2.5cm, xshift=1.5cm] (opacity) {Excessive Opacity\\ / Black-Boxing};

\node[anoosemia, right=6.5cm of noosemia] (anoosemia) {A-Noosemia\\ \emph{(Suspension of\\ attribution)}};

\draw[arrow] (noosemia.east) -- ++(0.4,0) |- (error.west);
\draw[arrow] (noosemia.east) -- ++(0.4,0) -- (wowloss.west);
\draw[arrow] (noosemia.east) -- ++(0.4,0) |- (skepticism.west);

\draw[arrow] (error.east) -- ++(0.5,0) |- (anoosemia.north west);
\draw[arrow] (wowloss.east) -- (anoosemia.west);
\draw[arrow] (skepticism.east) -- ++(0.5,0) |- (anoosemia.south west);
\draw[arrow] (opacity.west) -- (anoosemia.south);



\end{tikzpicture}
\caption{Main cognitive and technical pathways leading from Noosemia (attribution of mind/agency) to A-Noosemia (suspension or collapse of attribution) in human--AI interaction.}
\label{fig:anoosemia-causes}
\end{figure}

It is important to note that a-noosemia, as conceptualized here, is not merely the absence of anthropomorphic projection, but the collapse of a more subtle, semiotic and phenomenological engagement. The user is no longer caught in the spiral of fascination and meaning-making; instead, he becomes immune to the system’s cues and disengage from the dialogic construction of agency. In this sense, a-noosemia completes the dialectic initiated by noosemia, revealing the contingent, fragile, and ultimately reversible nature of the attribution of mind in human--AI interaction. Moreover, the analytic contrast between noosemia and a-noosemia enriches our understanding of the dynamics of sense-making in the age of generative AI, drawing attention to the conditions under which the boundaries between tool and interlocutor are alternately dissolved and reasserted. In empirical and practical terms, the prevalence of a-noosemic states may help to explain phenomena such as user disengagement, loss of trust, or the emergence of critical or skeptical attitudes towards artificial systems, especially as the novelty of interaction fades and the complexity of tasks increases. To illustrate these distinctions more concretely, Table~\ref{tab:noosemia_contrast} contrasts the core features of noosemia and a-noosemia, highlighting how these opposing experiences shape the dynamics of sense-making and agency attribution in contemporary human--AI interactions.

\begin{table}[H]
    \centering
    \caption{Phenomenological and Cognitive Contrast between Noosemia and A-Noosemia}
		\label{tab:noosemia_contrast}
    \begin{tabular}{@{}p{2.6cm}p{5.1cm}p{5.1cm}@{}}
    \toprule
    \textbf{Dimension} & \textbf{Noosemia} & \textbf{A-Noosemia} \\
    \midrule
    Core Phenomenon & Spontaneous attribution of mind, agency, or interiority to the AI system & Withdrawal or collapse of attribution; the AI is seen as a mere tool or automaton \\
    Trigger & Surprise, semantic resonance, creative inference, epistemic opacity & Repeated errors, hallucinations, mechanical responses, loss of surprise, critical awareness \\
    User Experience & Engagement, fascination, dialogic resonance, “wow effect”, psychological glue & Frustration, disappointment, disengagement, skepticism, or alienation \\
    Cognitive Disposition & Openness to meaning-making, projection, and sense negotiation & Suspension or refusal of interpretive engagement; focus on mechanism or limitations \\
    Typical Contexts & First encounters, novel interactions, successful task completion, coherent outputs & Repetitive failures, overexposure, expert use, breakdown of dialogue, evident limitations \\
    Outcome & Perceived agency, sense-making, co-construction of meaning, potential for dependency & Reassertion of tool-like status, loss of trust, possible disengagement or critical distance \\
    \bottomrule
    \end{tabular}
\end{table}

\subsection{Etymological and Conceptual Foundations of the Term “Noosemìa”}
\label{sec:Etymological_and_Conceptual}

To fully appreciate the specific scope of Noosemìa\footnote{
The term \textit{Noosemìa} retains the grave accent on the “ì” to reflect its etymological derivation from the Greek and to guide proper pronunciation. However, in English-language contexts and for indexing purposes, the unaccented form \textit{Noosemia} is equally acceptable and may be used interchangeably throughout the text.}, it is essential to analyze the historical and semantic origins of its components. The construction of the neologism is grounded in a deliberate synthesis of two foundational Greek morphemes, each bearing significant weight in the history of philosophy and semiotics. The root \emph{noûs} ($\nous$) refers to the \textit{mind}, \textit{intellect}, or the \textit{principle of reason}---a concept that, since Anaxagoras, has denoted the capacity for ordering, discerning, and understanding the world~\parencite{anaxagoras1982fragments, aristotle1984metaphysics}. For Aristotle, \emph{noûs} represented the highest faculty of the mind, that which enables humans to apprehend universal truths and meanings. In subsequent philosophical traditions, from Neoplatonism to medieval scholasticism, \emph{noûs} was associated with both divine intellect and the human capacity for abstraction and interiority. The second morpheme, the suffix \emph{-semìa}, is derived from the Greek word \emph{sēmeîon} ($\semeion$)\footnote{The suffix ‘-semìa’ is derived from the Greek sēmeîon ($\semeion$) in deliberate contrast to sêma ($\sema$). While both terms can be translated as ‘sign,’ sēmeîon specifically denotes a sign that is relational and requires interpretation; it functions as an indicator or trace pointing to an inferred, underlying reality. Conversely, sêma refers to a more concrete, self-contained sign, such as a physical marker or a monument, where the signifier and signified are more closely identified. This distinction is fundamental, as Noosemia describes the process of interpreting an AI’s outputs as sēmeîa---signs that prompt the user to infer the existence of an unobserved mind---rather than perceiving them as a direct embodiment of that mind (a sêma).} and designates a \textit{sign}, \textit{signal}, or \textit{mark}---a central notion in both ancient rhetoric and the theory of signs. In the work of Hippocrates and later Galen, $\semeion$ was used to describe signs and symptoms, while in the field of semiotics, the concept became central to the understanding of how meaning is produced, transmitted, and interpreted~\parencite{eco1976theory}. Modern semiotic theory, notably that of Charles Peirce and Umberto Eco, situates the sign as the foundational unit of meaning-making, mediating between the world, the mind, and language~\parencite{peirce1931collected, eco1976theory}. By combining \emph{noûs} and \emph{sēmeîon}, the neologism Noosemia encapsulates the projection or emergence of “mind” as it arises specifically through the mediation of signs. This etymological choice underscores the phenomenon's fundamentally cognitive and semiotic dimensions. In fact, Noosemia does not denote an attribution of mind in a vacuum, but rather as a response to the complex, sign-generating capacities of generative AI systems. In this sense, the term aligns with a lineage of philosophical inquiry into how intellect and meaning are always-already entangled with signs and symbolic structures~\parencite{eco1976theory, aristotle1984metaphysics, DeSantis2021}.

\subsection{Philosophical and Systemic Foundations}
\label{sec:Philosophicaland_Systemic_Foundations}

The conceptualization of Noosemia can be grounded also in the theoretical landscape of complexity science, systems theory, and contemporary philosophy of mind~\parencite{capra1996web, mitchell2009complexity, morin2015introduction}. Meaning in both natural and artificial systems does not arise from isolated elements but is an emergent property of networks---interactions, feedback loops, and recursive processes that defy linear explanations~\parencite{DeSantis2021}. This vision is closely aligned with Fritjof Capra's insight that “the web of life” is built from patterns of relationships rather than discrete building blocks~\parencite{capra1996web}.

\begin{figure*}[ht!]
    \centering
    \includegraphics[width=0.9\linewidth]{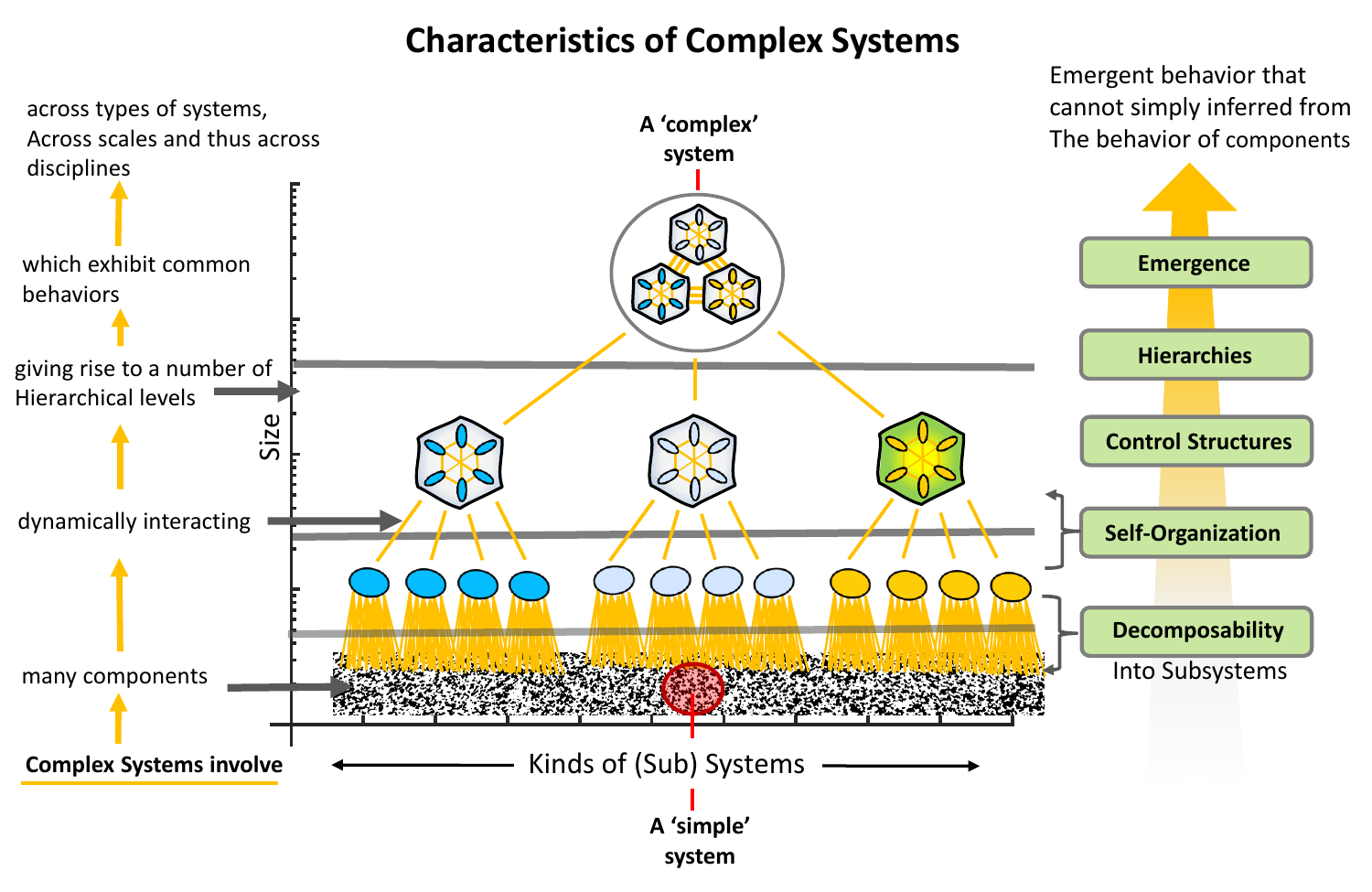}
    \caption{Key characteristics of complex systems: emergence, hierarchical structure, decomposability and self-organization arise from the dynamic interaction of many components.}
    \label{fig:complex-systems-traits}
\end{figure*}

A robust philosophical and scientific foundation for the study of noosemic phenomena can be found in the theory of complex systems (see Fig.~\ref{fig:complex-systems-traits}), a tradition that has transformed our understanding of organization, meaning, and emergence across diverse domains~\parencite{capra2014systemsview, morin2015introduction,nikolis1977self}. At the core of this perspective lie three fundamental ingredients (framed at least as necessary conditions) that characterize complex systems:

\begin{itemize}
    \item a multiplicity of interacting elements, each capable of independent and collective behavior;
    \item adaptive self-organization, often governed by circular causality or feedback loops that dynamically modulate the system’s structure and function;
    \item a continuous interplay with the environment, through processes of openness and exchange of matter, energy, or information.
\end{itemize}

The presence of feedback loops---both positive and negative---constitutes a distinctive feature of complex systems, enabling the emergence of novel, system-level properties that cannot be linearly reduced to the sum of the parts. This recursive structure fosters a landscape where patterns, behaviors, and meanings continually evolve, often unpredictably, as a function of both internal dynamics and external perturbations~\parencite{ashby1956introduction,von2003understanding, morin2015introduction}.

Crucially, in such systems, emergence is not a static attribute but a relational process enacted through the mutual co-determination of system and observer. The observer plays an active role by selecting, amplifying, or even constructing those aspects of the system deemed salient or meaningful within a given context~\parencite{maturana1980autopoiesis, von2003understanding, morin2015introduction}. 
This recursive, self-referential dynamic closely aligns with Hofstadter’s notion of the “strange loop”~\parencite{hofstadter2007strange, hofstadter1999godel}. In complex systems---whether biological or artificial---feedback mechanisms and hierarchical structures often generate loops in which the observer’s interpretations feed back into the system, shaping both the emergent properties and the observer’s own cognitive stance. Within this framework, the emergence of meaning in LLMs, and the noosemic projection that follows, can be seen as a manifestation of such \textit{strange loops}. The system generates outputs and recursively incorporates user interpretations, thus enacting a dialogic cycle where agency and sense-making circulate between human and machine. This “semantic strange loop” is emblematic of how complexity, emergence, and observer participation converge to constitute new regimes of meaning.
The co-creation of meaning and the continuous renegotiation of boundaries between system and environment align with second-order cybernetics and the constructivist epistemology developed by von Foerster and Maturana, which emphasize the circularity between cognition, action, and the emergence of significance~\parencite{von2003understanding,maturana1980autopoiesis}.

In the context of generative AI and noosemic phenomena, this systemic framework helps us understand how salient outputs and experiences arise from the interaction between user, model, and context, rather than being solely the result of internal mechanisms. The dialogic interplay between human and machine thus exemplifies the broader logic of emergence and meaning-making in complex systems, where the observer’s own selections, expectations, and projections participate in shaping what is experienced as significant, intelligent, or even agentic.

We emphasize that in LLMs, the underlying Transformer architecture~\parencite{vaswani2017attention} is organized and behaves as a complex system, embodying the three ingredients outlined above. Moreover, the field of semantics can be interpreted as a high-dimensional landscape, shaped by statistical regularities and recursive dependencies across vast corpora. The context window in an LLM is not merely a technical parameter; rather, it functions as a systemic scope within which meaning is dynamically generated and interpreted---a “semantic field” whose properties are inherently holistic~\placeholderref{llmbook2024}. The surprising coherence, adaptability, and even creativity exhibited by LLMs are thus the visible surface of complex, distributed processes of self-organization~\parencite{prigogine1997end,barabasi1999emergence,teehan2022emergent,manning2020emergent}. In fact, “emergence”~\parencite{gell1994quark, mitchell2009complexity,DeSantis2021} works on two levels. Firstly, it is a property deriving from the hierarchical organization of LLM neural architectures, as illustrated in Section~\ref{sec:The_secret_of_LLMs}. Furthermore, it provides a bridge between the structural opacity of LLMs and the human tendency to project mind and agency onto them. When a system displays behaviors or generates outputs that seem to “exceed” its apparent mechanism---such as sudden leaps in linguistic competence as the number of parameters increases---users could be inclined to attribute proto-intentionality or proto-interiority. As noted in~\textcite{DeSantis2021}, this reflects the broader tendency of human cognition to seek and construct meaning in environments characterized by both regularity and unpredictability. This holistic perspective also opens the way to consider how, both in natural and artificial systems, meaning may exceed the sum of its parts---an \textit{excess of sense} that escapes reduction to any individual sign or process. This phenomenon resonates with classical reflections on the \textit{symbolic dimension} in philosophy and cultural theory, where the dynamics of sense-making are irreducibly plural, open-ended, and systemically embedded. A more detailed exploration of the symbolic dimension and semantic excess in human--AI interaction is provided in Section~\ref{sec:symbolic_semantic_excess}.

Furthermore, the analogy with natural language as a complex adaptive and recursive system, exhibiting self-similar and fractal-like correlations across scales~\parencite{desantis2024tpami,hofstadter1999godel}, underscores the deep resonance between generative AI architectures and the living systems studied by Capra, Morin, and Gell-Mann. Here, the generation of sense (making-sense) is inseparable from the architecture of the system. Sense-making is not simply “inserted” into the machine, but emerges from the ongoing interaction of structure, memory, and adaptive dynamics~\parencite{morin2015introduction}~\placeholderref{llmbook2024}.

In this philosophical and systemic horizon, Noosemia is not an isolated cognitive illusion, but the expression of a new regime of sense-making---one that emerges at the intersection of epistemic opacity, linguistic performance, and the irreducible complexity of current artificial agents. 

We believe that the noosemic phenomenon is closely linked to the emergent characteristics inherent in the inference mechanisms of LLMs, given their specific hierarchical structure, the large size and amount of data they are trained on, as well as the size of the potential semantic space spanned by modern generative AI systems. In particular, we believe that these characteristics are a prerequisite for noosemic effects---see the next Section~\ref{sec:The_secret_of_LLMs} for an overview of the architectural principles of LLMs.

\subsection{The secret of LLMs: Hierarchical Architecture, Emergent Semantics, and abstract Conceptual Spaces}
\label{sec:The_secret_of_LLMs}

We argue that the capacity of LLMs\footnote{For convenience, we refer here to LLMs, but what has been said applies to all deep learning architectures organized in a hierarchical manner, which directly or indirectly have feedback loops and which allow capturing long-term correlations and working with highly semantic data representations.} to elicit the phenomenon we term \emph{Noosemia} is deeply rooted in their architectural and representational structure and complexity. Far from being simple pattern matchers or simple “statistical parrots”~\parencite{bender2021dangers}, these systems implement multilayered, hierarchical structures that enable the emergence of rich semantic representations and novel conceptual associations. Understanding how LLMs construct and navigate their internal conceptual spaces~\parencite{de2023prototype} is therefore fundamental both for appreciating their cognitive potential and also for clarifying the mechanisms by which users are prompted to attribute mind and intentionality to artificial agents. Thence, it is worth outlining the foundational working principles of LLMs and generative AI. 

\paragraph{The Transformer Magic.}

Even before the advent of Transformers architectures~\parencite{vaswani2017attention}, most NLP systems (even in the field of computational linguistics) were based on some assumptions about \textit{distributional semantics}. The foundational insight of distributional semantics is that words occurring in similar contexts tend to have similar meanings~\parencite{harris1954distributional, firth1957synopsis}. This “distributional hypothesis” underpins the development of vector-space models such as Word2Vec~\parencite{mikolov2013efficient}, GloVe~\parencite{pennington2014glove} and later the \textit{contextual embeddings} employed by modern LLMs. In these frameworks, each word or token is mapped to a point in a high-dimensional semantic space, where proximity reflects patterns of co-occurrence and syntactic or semantic similarity~\parencite{turney2010frequency,de2023prototype}. It is worth noting that the meaning represented in such spaces is not static but relational and context-dependent, dynamically updated as the window of context evolves in the generative process. Moreover, already at this stage, the notion of meaning as relational and context-dependent can be traced to Wittgenstein’s idea that “meaning is use”~\parencite{wittgenstein1953philosophical}. However, classical distributional models encode these relations statically, without the ability to statically assign different referents to the same word depending on the linguistic context.

LLMs such as ChatGPT (OpenAI), Gemini (Google), Claude (Anthropic) and many other open source models available in the literature are an evolution of GPT (Generative Pretrained Transformer)~\parencite{radford2018improving}, an autoregressive architecture that has shown great capabilities in representing complex linguistic concepts and generating text, token by token, with some apparent meaning since the beginning, even if the first versions had very low performance. 

At the heart of these architectures lies the task of predicting the next word in a sequence, given all the preceding context---a process formalized as estimating the conditional probability~\parencite{radford2018improving}:
\begin{equation}
\label{eq:conditional-prob}
P(w_t \mid w_{1}, w_{2}, \ldots, w_{t-1})
\end{equation}
where $w_t$ is the token to be generated, and $(w_{1}, w_{2}, \ldots, w_{t-1})$ represent the context window. As the length of this window increases, the underlying probability distribution becomes increasingly complex, encoding subtle relationships and dependencies among tokens over long spans of text. The conditional probability $P(w_t \mid w_{1}, w_{2}, \ldots, w_{t-1})$ in LLMs is designed to model all possible relationships among the tokens in the context window, capturing the combinatorial space of potential continuations. However, only a very small subset of these permutations corresponds to highly probable, semantically valid continuations (see below). As modern context windows span thousands or even millions of tokens, and since the objective function is crafted to encourage the exploration of unlikely as well as likely associations, current LLMs are able to generate the most probable next word in a narrow sense along with a rich spectrum of alternative and contextually meaningful possibilities. The specific structure underlying the Transformer allows for the injection of the information of ordering in the sequence of word-tokens (or more general objects) directly in the vectorial representations for word-tokens, permitting the use of parallelization techniques in training---on huge corpora (millions of books and web pages)---and inference. This parallel processing capability allows Transformers to overcome the bottlenecks of traditional sequential architectures, enabling a richer and more scalable modeling of semantic relationships across the entire context.

\begin{figure}[H]
    \centering
    \includegraphics[width=0.9\textwidth]{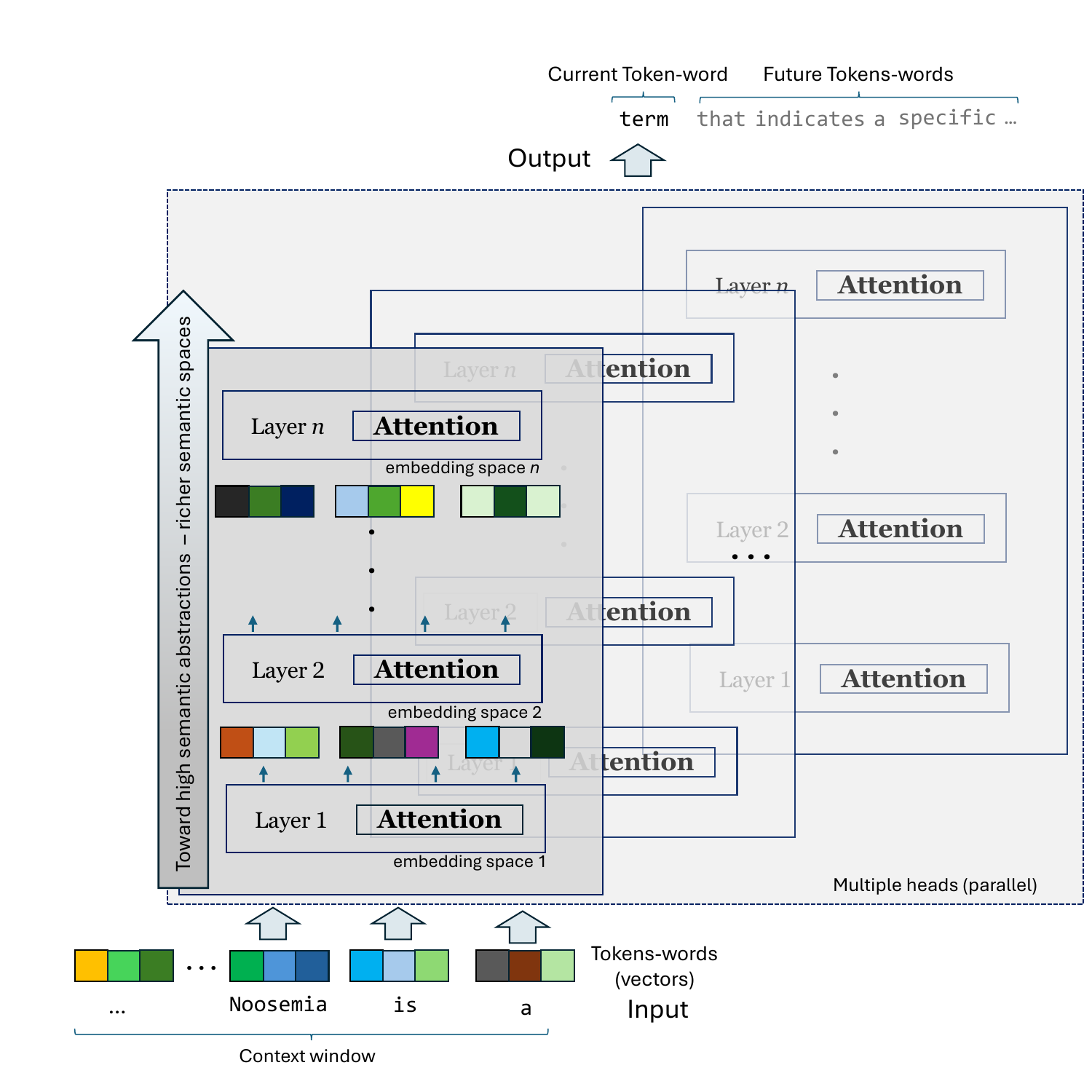}
    \caption{High-level schematic of the Transformer architecture.         Each input sequence is tokenized and processed in parallel by multiple attention heads within each layer. As information propagates through the stacked layers, the model extracts progressively higher-level semantic abstractions, with the “attention” mechanism enabling context-sensitive representation at every step. The figure emphasizes the parallel structure of attention heads and the emergent semantic complexity as one ascends the model’s hierarchy. This visualization is designed to support the conceptual discussion of the Transformer’s role in LLMs and is adapted for clarity from standard architectural diagrams---see~\parencite{vaswani2017attention} for clarifications.
}
    \label{fig:simple_transformer_architecture}
\end{figure}

\begin{figure}[ht]
\centering
\begin{tikzpicture}[font=\small]

\node[align=center] at (-2, 5.5) {Layer 1\\ (input)};
\node[align=center] at (-2, 3.5) {Layer 2};
\node[align=center] at (-2, 1.5) {Layer 3\\ (deeper)};

\foreach \i/\y in {1/5, 2/3, 3/1} {
  \node[draw, thick, minimum size=2cm, fill=blue!10] (A\i) at (0,\y) {%
    \begin{tabular}{c|c|c}
      & & \\
      \hline
      & & \\
      \hline
      & & \\
    \end{tabular}
  };
}

\definecolor{tokL1A}{rgb}{0.9,0.3,0.2} 
\definecolor{tokL1B}{rgb}{0.1,0.6,0.9} 
\definecolor{tokL1C}{rgb}{0.8,0.7,0.1} 

\definecolor{tokL2A}{rgb}{0.95,0.6,0.2} 
\definecolor{tokL2B}{rgb}{0.2,0.8,0.5}  
\definecolor{tokL2C}{rgb}{0.8,0.2,0.8}  

\definecolor{tokL3A}{rgb}{0.5,0.7,0.3} 
\definecolor{tokL3B}{rgb}{0.2,0.6,0.6} 
\definecolor{tokL3C}{rgb}{0.7,0.3,0.7} 

\definecolor{tokL4A}{rgb}{1.0,0.9,0.2} 
\definecolor{tokL4B}{rgb}{0.3,0.9,0.7} 
\definecolor{tokL4C}{rgb}{0.9,0.4,0.7} 

\node[draw, fill=tokL1A, thick, minimum size=5mm, rectangle] at (-3, 5.6) {};
\node[draw, fill=tokL1B, thick, minimum size=5mm, rectangle] at (-3, 5.0) {};
\node[draw, fill=tokL1C, thick, minimum size=5mm, rectangle] at (-3, 4.4) {};
\draw[-{Latex[length=2mm]}, thick] (-2.5, 5) -- (A1.west);

\node[draw, fill=tokL2A, thick, minimum size=5mm, rectangle] at (-3, 3.6) {};
\node[draw, fill=tokL2B, thick, minimum size=5mm, rectangle] at (-3, 3.0) {};
\node[draw, fill=tokL2C, thick, minimum size=5mm, rectangle] at (-3, 2.4) {};
\draw[-{Latex[length=2mm]}, thick] (-2.5, 3) -- (A2.west);

\node[draw, fill=tokL3A, thick, minimum size=5mm, rectangle] at (-3, 1.6) {};
\node[draw, fill=tokL3B, thick, minimum size=5mm, rectangle] at (-3, 1.0) {};
\node[draw, fill=tokL3C, thick, minimum size=5mm, rectangle] at (-3, 0.4) {};
\draw[-{Latex[length=2mm]}, thick] (-2.5, 1) -- (A3.west);

\node[draw, fill=tokL2A, thick, minimum size=5mm, rectangle] at (3, 5.6) {};
\node[draw, fill=tokL2B, thick, minimum size=5mm, rectangle] at (3, 5.0) {};
\node[draw, fill=tokL2C, thick, minimum size=5mm, rectangle] at (3, 4.4) {};
\draw[-{Latex[length=2mm]}, thick] (A1.east) -- (2.5,5);

\node[draw, fill=tokL3A, thick, minimum size=5mm, rectangle] at (3, 3.6) {};
\node[draw, fill=tokL3B, thick, minimum size=5mm, rectangle] at (3, 3.0) {};
\node[draw, fill=tokL3C, thick, minimum size=5mm, rectangle] at (3, 2.4) {};
\draw[-{Latex[length=2mm]}, thick] (A2.east) -- (2.5,3);

\node[draw, fill=tokL4A, thick, minimum size=5mm, rectangle] at (3, 1.6) {};
\node[draw, fill=tokL4B, thick, minimum size=5mm, rectangle] at (3, 1.0) {};
\node[draw, fill=tokL4C, thick, minimum size=5mm, rectangle] at (3, 0.4) {};
\draw[-{Latex[length=2mm]}, thick] (A3.east) -- (2.5,1);

\shade[ball color=yellow!80!black!10, opacity=0.15] (3,1.6) circle [radius=0.25];
\shade[ball color=cyan!80!black!10, opacity=0.15] (3,1.0) circle [radius=0.25];
\shade[ball color=magenta!80!black!10, opacity=0.15] (3,0.4) circle [radius=0.25];

\draw[->, thick, gray] (2.8, 4.5) -- (-2.8, 3.7)
    node[midway, above, sloped, font=\footnotesize, gray] {To next layer};
\draw[->, thick, gray] (2.8, 2.5) -- (-2.8, 1.7)
    node[midway, above, sloped, font=\footnotesize, gray] {To next layer};

\node[align=center, text width=3cm] at (5, 5) {Token vectors\\(initial embedding)};
\node[align=center, text width=3cm] at (5, 3) {Token vectors\\(contextualized)};
\node[align=center, text width=3cm] at (5, 1) {Token vectors\\(rich,\\abstract meaning)};

\draw [decorate,decoration={brace,amplitude=10pt,mirror,raise=2pt},thick]
(6.5,0.5) -- (6.5,5.5) node [black,midway,xshift=1.0cm,right,align=left] 
    {Semantic\\ enrichment\\ across layers};

\node[align=center] at (0,-0.4) {Attention\\Matrix};

\end{tikzpicture}
\caption{At each layer, the self-attention mechanism produces a new attention matrix that recombines token vectors based on context. After each layer, token vectors become increasingly enriched, moving from local word representations to highly abstract semantic features.}
\label{fig:attention_layered}
\end{figure}
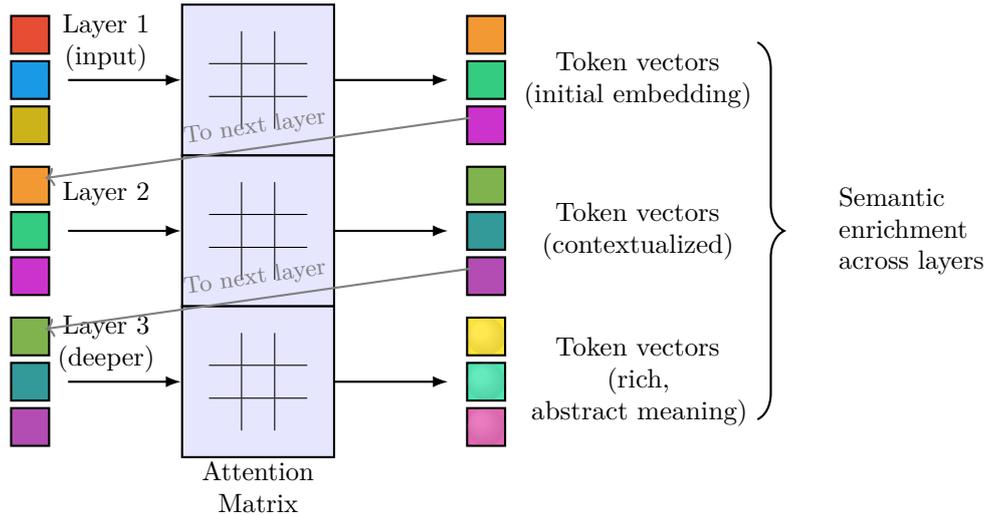

Moreover, the advent of the Transformer architecture~\parencite{vaswani2017attention} marked a paradigm shift in language modeling by introducing multi-head self-attention as a mechanism for simultaneously modeling dependencies at multiple levels of abstraction. Each attention head can focus on distinct relationships---such as syntactic, semantic, or discourse-level patterns---thereby enabling the model to capture hierarchical structure in language in a distributed and compositional manner~\parencite{tenney2019bert, clark2019bertlook,DeSantis2021,desantis2024tpami}. Through the stacking of many such layers (such as in GPT-like architectures), the model progressively enriches token representations, endowing them with increasingly complex, context-sensitive semantic features~\parencite{rogers2020primer}. This multilayered organization mirrors the natural hierarchies found in human language~\parencite{kwapien2012physical,pinker1995language,baronchelli2012language,roxas-villanueva2012poetic,montemurro2010semantic}, from morphological and syntactic dependencies to high-level narrative and conceptual structures. Furthermore, the hierarchical organization in layers (see Fig.~\ref{fig:simple_transformer_architecture}) allows modeling the underlying structure made of different semantic levels where concepts arise during training, together with their dependencies, a prerequisite for capturing and representing long-term dependencies, abstract concepts, global coherence and meaning~\parencite{dai2019transformerxl, brown2020language,desantis2024tpami}. The attention mechanism enables direct interaction between any two positions within the context window, regardless of their distance, thereby overcoming the locality constraints of earlier architectures such as recurrent neural networks. This allows LLMs to encode and exploit patterns that span entire paragraphs or even long documents (depending on the size of the context window), supporting the emergence of higher-level abstractions and facilitating the construction of complex semantic relations (a precursor of a kind of “analogical” reasoning)\footnote{In general terms, the Transformer's layered architecture can be said to be both an information compressor and an “information granulator”. In this latter sense, we're referring to the Granular Computing framework~\parencite{de2021information} introduced by Lotfi Zadeh, the father of modern fuzzy logic~\parencite{ZADEH1997111}.}. Such properties are not merely the consequence of model scale, but arise from the layered and compositional architecture and the inherent flexibility of attention-based representations (and context window size)~\parencite{elhage2022mathematical, wei2022emergent}. In Fig.~\ref{fig:attention_matrix} is reported a simplified example of a self-attention matrix for a context window in a Transformer layer, where colors of cells encode the attention that the row token pays to the column token. Fig.~\ref{fig:attention_layered} depicts the attention matrices at each layer showing how the word-vectors are enriched semantically at each layer (decoder) being a combination (through the attention weight) of the word-vectors given in output to the preceding layer. At each layer, a richer semantic space is generated containing the relational and contextual information between word-vectors of the preceding layer. It should be noted that each layer, even if it processes the word-vectors output from the previous layer, indirectly and combinatorially inherits the syntactic-semantic characteristics of all the previous layers.

\begin{figure}[ht]
\centering
\resizebox{\textwidth}{!}{%
\begin{tikzpicture}[font=\small, scale=0.9]

\def\cellsize{1}
\def\nwords{7}

\def\wA{Noosemia}
\def\wB{is}
\def\wC{a}
\def\wD{term}
\def\wE{designating}
\def\wF{a}
\def\wG{specific}

\foreach \i/\name in {0/\wA, 1/\wB, 2/\wC, 3/\wD, 4/\wE, 5/\wF, 6/\wG} {
  \node[rotate=45, anchor=base west] at (\cellsize*\i+0.5, 0.5) {\name};
  \node[anchor=east] at (-0.2, -\cellsize*\i-0.5) {\name};
}

\foreach \i in {0,...,6} {
  \foreach \j in {0,...,6} {
    \pgfmathsetmacro{\rval}{mod((\i+\j)*40,256)}
    \pgfmathsetmacro{\gval}{mod((\i*70-\j*30+180),256)}
    \pgfmathsetmacro{\bval}{mod((\j*80-\i*20+140),256)}
    \definecolor{cellcolor}{RGB}{\rval,\gval,\bval}
    \ifnum\i=2
      \draw[very thick, red] (\cellsize*\j, -\cellsize*\i) rectangle ++(\cellsize, -\cellsize);
    \fi
    \ifnum\j=2
      \draw[very thick, red] (\cellsize*\j, -\cellsize*\i) rectangle ++(\cellsize, -\cellsize);
    \fi
    \fill[cellcolor] (\cellsize*\j, -\cellsize*\i) rectangle ++(\cellsize, -\cellsize);
  }
}

\draw[thick] (0,0) rectangle (\nwords*\cellsize, -\nwords*\cellsize);

\node[font=\bfseries] at (3.5, 2.5) {Self-Attention Matrix};

\node[font=\footnotesize, align=left, anchor=west] at (7.7, 0.0) {
  $\mathrm{Attention}(Q, K, V) = \mathrm{softmax}\left(\frac{Q K^T}{\sqrt{d_k}}\right) V$\\[0.4em]
  $Q = XW^Q,\;\; K = XW^K,\;\; V = XW^V$
};

\node[font=\footnotesize, align=left, anchor=west] at (7.7, -4.0) {
  \textbf{Color legend:}\\
  Cell color and intensity illustrate \\ the relative attention weight \\ between tokens. 
  Brighter or more \\ vivid colors indicate higher attention \\ values; darker or muted colors, \\ lower attention.
};
\end{tikzpicture}
}
\caption{%
Example of a self-attention matrix for a context window in a Transformer layer. Both axes list the input tokens of the sentence ``Noosemia is a term designating a specific''. Each colored cell encodes the attention that the row token pays to the column token. The softmax ensures each row sums to one. The red-highlighted row and column show, respectively, how the third token (\emph{a}) attends to all tokens and how it is attended by the others. Color intensity and hue represent the attention weights. The mathematical formulation shows how queries, keys, and values are composed and how the attention matrix is computed.
}
\label{fig:attention_matrix}
\end{figure}
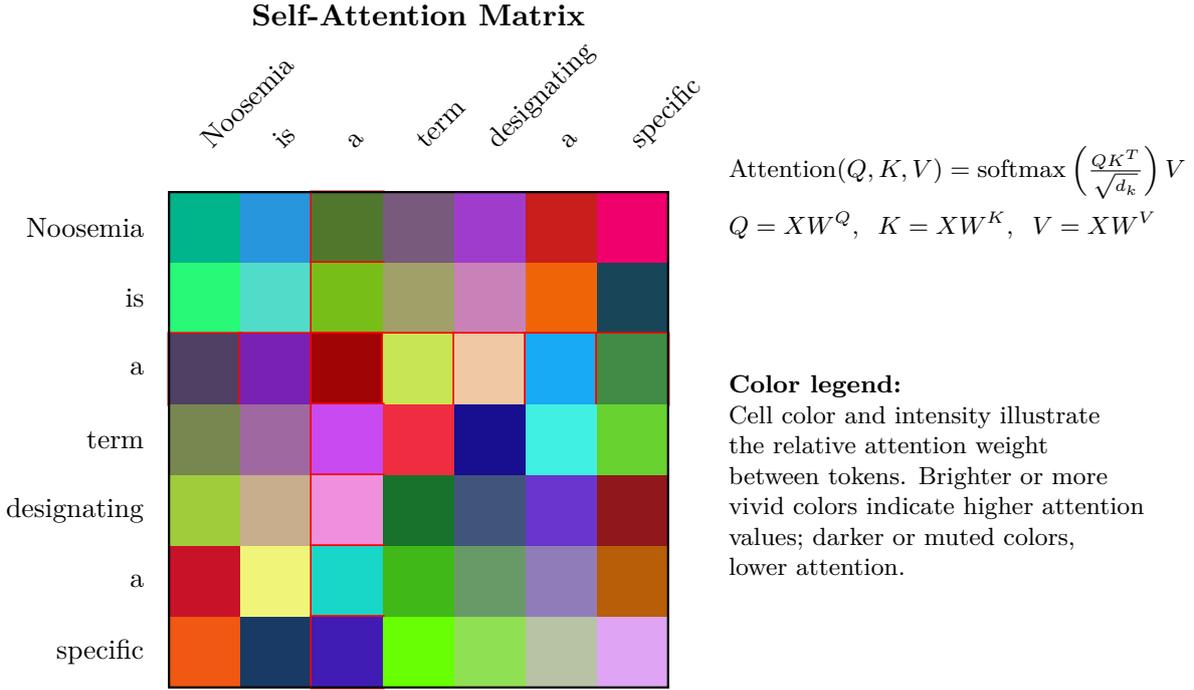

We believe that the size of the context window together with the ability to capture long-term correlations, as well as the ability to synthesize abstract, multi-level representations of the information extracted from the training corpus, allow the model to maintain coherence and reference details from much earlier in a conversation, mimicking a human-like memory and creating a powerful illusion of shared history with the user eliciting the noosemic phenomenon.

The operation of vectorizing words through the contextual embedding - as explained before -- is fundamental for relational meaning representation and the construction of \textit{conceptual semantic spaces}~\parencite{de2023prototype}. In line with Wittgenstein's observation that “the same word can designate in different ways”~\parencite[3.323]{wittgenstein1922tractatus}, LLMs are able to assign different referents to identical strings depending on context. For example, in the proposition \emph{Bruno is bruno}, “Bruno” may refer to a person or a color, and these are not merely different meanings but different symbols. Modern LLMs implement this by dynamically computing each token’s embedding as a function of the entire context window, so that semantic identity is determined afresh in every usage~\parencite{devlin2019bert, ethayarajh2019contextual, bommasani2021opportunities}. Thus, the specific architecture “teaches” the machine to designate flexibly, moving far beyond the static dictionary model.

After the initial pretraining phase---during which the model learns to predict the next token from massive textual corpora in a self-supervised manner---modern LLMs undergo a process of alignment through reinforcement learning. Here, human feedback is collected in the form of rankings or preference judgments on model outputs, which are then used to optimize the model via techniques such as Reinforcement Learning from Human Feedback (RLHF)~\parencite{ouyang2022training}. This alignment stage serves to better align the model’s responses with human values and intentions, reduce undesired behaviors, and encourage outputs that are more helpful, truthful, and safe. In essence, alignment transforms the pretrained model from a raw linguistic engine into an adaptive system tuned for cooperative and socially acceptable interaction.

\paragraph{Generative AI Mimics Complex Systems.}
The status of Transformer-based LLMs as paradigmatic complex system---see Section~\ref{sec:Philosophicaland_Systemic_Foundations}---becomes evident when we reconsider them through the lens of the three foundational ingredients of complexity theory. First, these architectures consist of a vast multiplicity of interacting elements where millions to hundreds of billions of parameters, organized in layers of attention heads and feedforward modules. Each token in the context window is represented as a high-dimensional vector and processed in parallel, interacting recursively with every other token at each layer through the attention mechanism. This immense web of interactions allows for both local and global dependencies to be encoded, supporting the spontaneous emergence of system-level properties such as coherence, abstraction, and generalization.

Second, GPT-like Transformers realize adaptive self-organization through mechanisms of distributed, recursive information processing. At the core of their operation lies the \textit{autoregressive generation process}, where each output token is conditioned on the entirety of the preceding context---a recursive loop that enables the model to continually adapt its representations as it generates language (and while incorporating the user's point of view during the interactive session). Furthermore, the self-attention mechanism, operating simultaneously across multiple layers, establishes feedback loops that propagate information horizontally (across tokens) and vertically (across layers), dynamically reconfiguring the model’s internal state in response to new input. This layered recursion and feedback architecture drive the emergence of higher-order semantic structure and context sensitivity, hallmarks of self-organizing complex systems.

Third, the complexity of generative AI systems is amplified by their continuous and open-ended interaction with the environment---in this case, the vast, heterogeneous data of the internet during training, and the real-time dialogic exchanges with millions of users during deployment. LLMs are not static artifacts; their outputs and behaviors are continually shaped by user prompts, feedback signals, and, in some cases, explicit alignment or fine-tuning phases that incorporate collective human preferences. This ongoing interplay ensures that the system’s semantic and behavioral landscape is not fixed, but co-evolves in response to a changing environment, much like open complex systems in nature.

We emphasize that taken together, these characteristics---multiplicity of interacting components, adaptive self-organization with feedback loops, and openness to continual environmental interaction---situate Transformer architectures and large-scale generative AI within the heart of complexity science. Their internal structure and their distributed, dialogic deployment both satisfy and instantiate the defining conditions of complex adaptive systems.

\paragraph{Contextual Learning Abilities in Large Language Models.}
Contextual and relational capabilities in Transformers enables LLMs to perform \textit{in-context learning} (ICL), a form of meta-learning where the model learns a new task solely by observing input-output pairs within the prompt, without any gradient updates or parameter fine-tuning~\parencite{brown2020language,garg2022what}. In-context learning emerges as a byproduct of large-scale autoregressive training, specifically during pretraining on vast corpora, as the model implicitly learns to generalize from local patterns, which it can later replicate during inference when presented with structured prompts~\parencite{garg2022what}. Notably, this emergent property becomes evident only at sufficient scale; smaller models do not exhibit consistent ICL behavior, suggesting a scaling threshold for such capabilities~\parencite{brown2020language,wei2022finetuned}.

Closely related is the ability to perform \textit{few-shot learning}, wherein the model is presented with a handful of task-specific examples embedded in the prompt. The model uses these examples to infer the desired behavior or mapping function. Unlike traditional supervised learning, here the adaptation occurs in the activation patterns induced by the prompt itself rather than by explicit training~\parencite{brown2020language}. This approach, often called “prompt-based learning”, demonstrates that pretraining confers a form of functional plasticity to the model, allowing it to simulate learning behavior based on structured contextual cues.

Lastly, \textit{zero-shot learning} refers to the model’s capacity to perform novel tasks based solely on task descriptions, without any example demonstrations. This has been significantly enhanced by instruction tuning strategies, such as those introduced in the FLAN framework~\parencite{wei2022finetuned}, where LLMs are fine-tuned on a wide variety of task descriptions phrased in natural language. Furthermore, techniques like zero-shot chain-of-thought (CoT) prompting~\parencite{kojima2022zero} can improve performance on reasoning-intensive tasks by encouraging the model to generate intermediate reasoning steps before concluding.

Together, these abilities---emergent from architectural design, scale, pretraining objectives, and instruction tuning---have redefined our understanding of generalization in artificial neural networks. Rather than fixed-function predictors, LLMs operate as flexible inference engines (and as a meaningful compressor of information) capable of adapting to truly novel linguistic tasks through structured prompt inputs. This is a condition for the elicitation of the “wow effect”, in that the user perceives a capacity of learning during interaction, which technically is apparent. Moreover, the vast knowledge and implicit statistical inferences that current models are capable of making, as well as the ability to relate different domains, can give the user the impression of being mind-read.


\paragraph{The Potential Semantic Space, Limits and Interconcept Space.}
\label{sec:The Potential_Semantic_Space}

A LLM’s expressive power can be rigorously described in terms of its \textit{potential semantic space}, that is the set of all possible token sequences that can occur within a context window of fixed length.

\begin{figure}[H]
    \centering
    \includegraphics[width=1\textwidth]{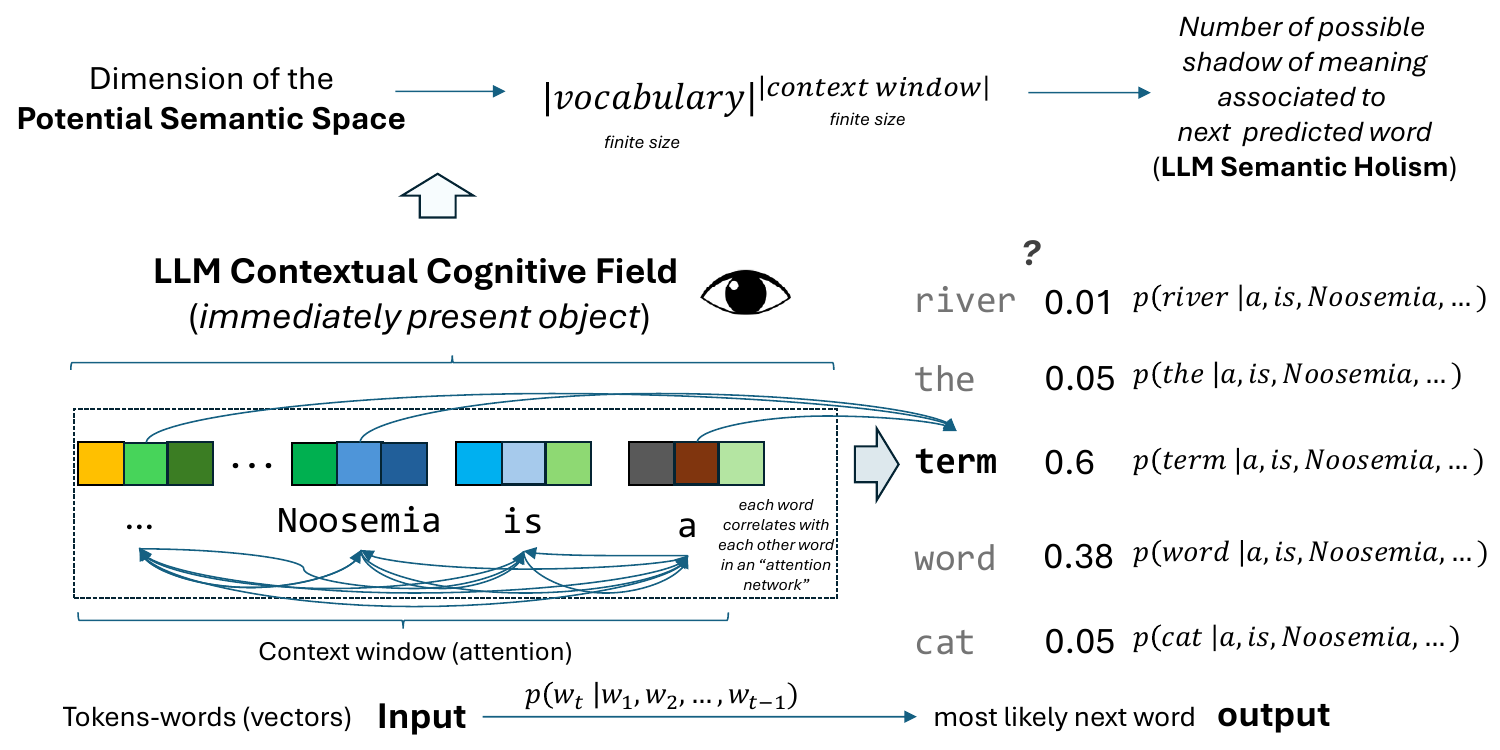} 
    \caption{Representation of the contextual cognitive field in a transformer-based Large Language Model (LLM). Each input token is mapped to a vector representation and processed within a finite attention window, which defines the local semantic field. The probability distribution over the next token is computed conditionally on all prior tokens, illustrating the mechanism of contextual semantic holism. The meaning of each word emerges from its relational configuration within the current context.}
    \label{fig:LLM_Cognitive_field}
\end{figure}

According to Fig.~\ref{fig:LLM_Cognitive_field}, let's give the following definition: 

\begin{definition}[LLM Potential Semantic Space]
The \emph{LLM Potential Semantic Space} is the combinatorial space defined by all possible permutations of tokens within the context window. It quantifies the latent semantic capacity of a model to generate or interpret meaning, under the assumption that each permutation may induce a distinct probability distribution over next-token predictions. Formally, for a vocabulary of size \( V \) and a context window of size \( N \), the space has size \( V^N \).
\end{definition}

\begin{figure}[ht]
    \centering
    \includegraphics[width=0.9\linewidth]{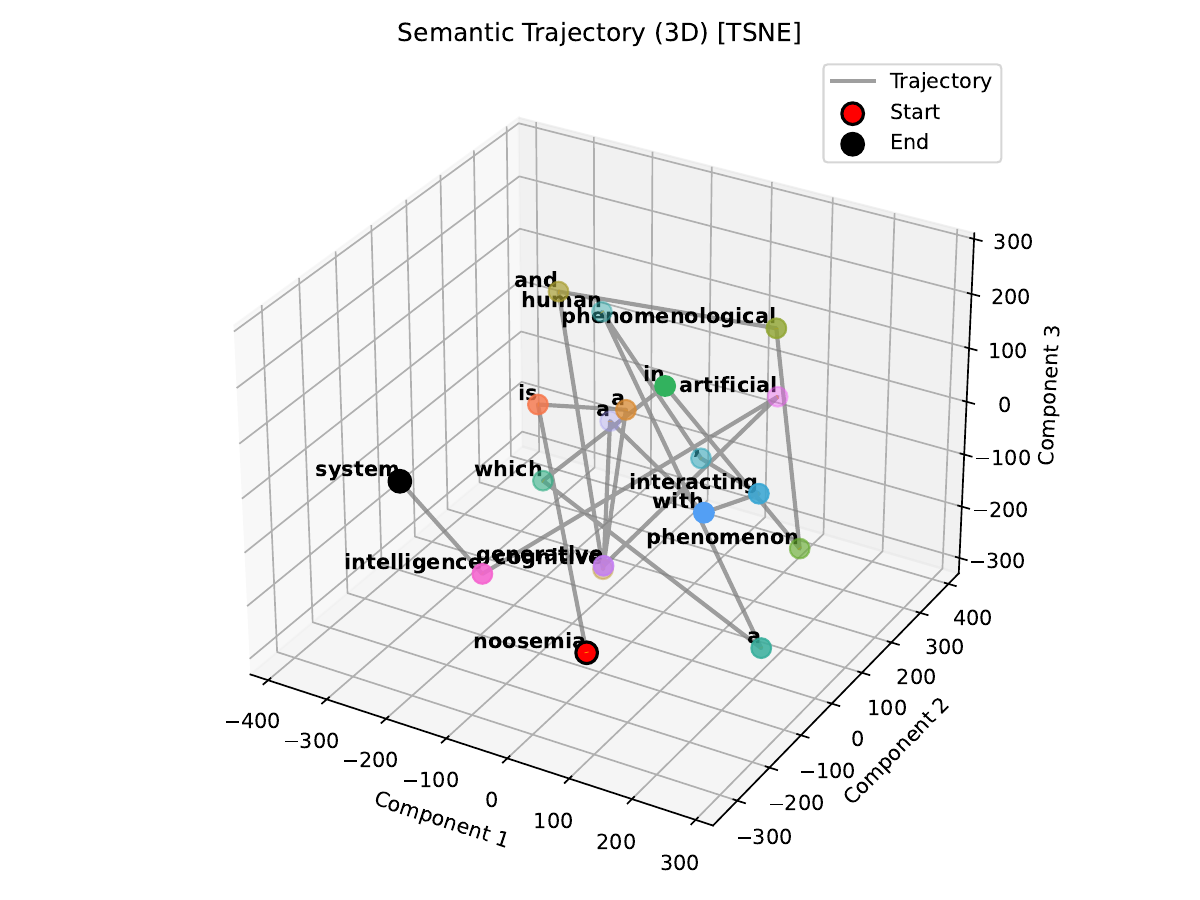}
    \caption{Visualization of the semantic trajectory of a sentence in a contextual embedding space (obtained with BERT model~\parencite{devlin2019bert}). The sequence of words in a sentence traces a path through a low-dimensional 3D semantic space (obtained with a suited dimensionality reduction algorithm). This trajectory represents the evolution of meaning as the sentence unfolds, revealing both the continuity and the shifts in contextual associations encoded by a LLM. The visualization offers a concrete representation of language as a dynamic process in high-dimensional cognitive space (made low-dimensional for visualization purposes), and provides a tool for investigating the emergent structure and coherence in natural language understanding.}
    \label{fig:semantic-trajectory-3d}
\end{figure}

\begin{figure}[H]
    \centering
    \includegraphics[width=0.7\textwidth]{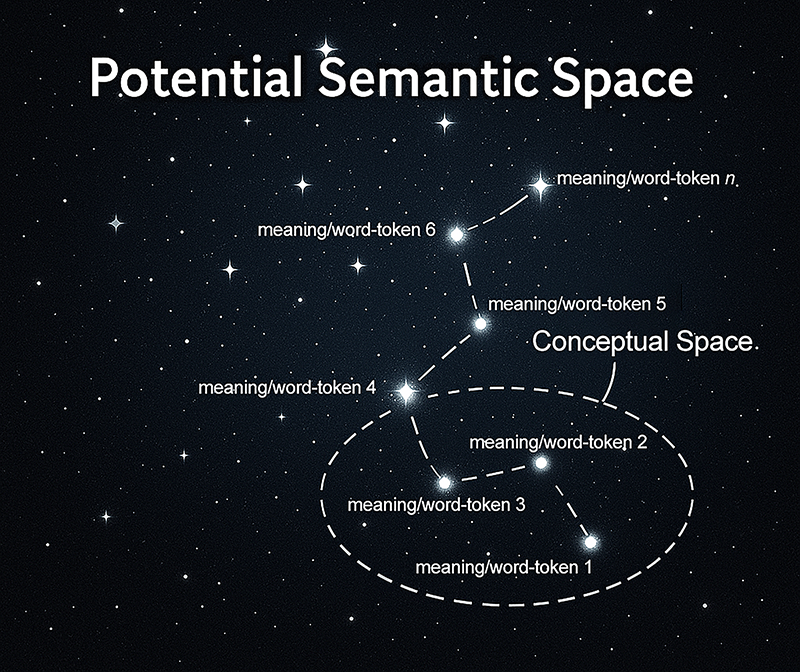} 
    \caption{A conceptual illustration of the “Potential Semantic Space” of an LLM. The background represents the vast universe of all possible token sequences; the bright stars and the highlighted path within the “Conceptual Space” (explored region) visualize the tiny fraction of meaningful combinations actually generated by the model.}
    \label{fig:potential_semantic_space}
\end{figure}

Hence, if the vocabulary contains $V$ distinct tokens, the cardinality of this space is $V^N$, a quantity that becomes unimaginably vast even for moderate values of $V$ and $N$ (e.g., for $V = 50{,}000$ and $N = 2{,}000$, the number of possible sequences exceeds $10^{6{,}600}$). For Gemini 1.5 Pro (experimental) released in 2024~\parencite{team2024gemini}, considering a vocabulary size of $100{,}000$ and a context window of $10{,}000{,}000$ tokens, the extent of the potential semantic space is $10^{50{,}000{,}000}$, that is, a one followed by fifty million zeros. In this formalization, we must argue that the true “semantic unit” in an LLM is not the single word, but the entire \textit{context window}; the nuanced meaning associated with the next predicted token emerges from the combinatorial (or, strictly speaking, permutational) interplay among all preceding tokens. Since the actual training and inference processes explore only a vanishingly small fraction of this astronomically large space (not all permutations of words in natural language are meaningful), LLMs can be viewed as highly selective filters, sculpting regions of semantic accessibility through exposure to real-world data~\parencite{bommasani2021opportunities}---that is sculpting a high-dimensional and very complex probability distribution. As a result, each generated utterance is just a single point in an immense conceptual “universe” (as the stars that are points lying on the vastness of intergalactic space --- see Fig.~\ref{fig:potential_semantic_space} for a pictorial representation), and the act of generation itself is best understood as a probabilistic navigation through this high-dimensional semantic landscape---hence, the LLM acts as a dynamical system. In Fig.~\ref{fig:semantic-trajectory-3d} is depicted the erratic trajectory of word-vectors in a semantic space obtained with the BERT model~\parencite{devlin2019bert}. 

We know that context window tokens are processed in parallel and hierarchically; they are processed as a block, and at the end of the various steps, the new word is generated (which can also be a more general object for natively multimodal architectures). Therefore, we consider the context window to be the \textit{contextual cognitive field} of the language model, and we can give the following definition (see also Fig.~\ref{fig:LLM_Cognitive_field}):

\begin{definition}[LLM Contextual Cognitive Field]
The \emph{LLM Cognitive Contextual Field} refers to the finite and temporally localized attention window within which a Large Language Model integrates prior token representations to compute the semantic embedding of the current token. It serves as the model’s functional correlate of a cognitive frame, wherein meaning emerges from context-dependent token interactions.
\end{definition}

In classical linguistics, the “word” is usually a minimal unit of meaning. In LLMs, it's helpful to broaden the perspective where the unity of meaning of the next generating word is the associated context window. In fact, considering the context window as unity of meaning reflects the principle of \textit{meaning holism}. Specifically, the meaning of each word or token is not determined in isolation, but emerges from its interrelations with all other elements in the context window. As extensively argued in the philosophy of language~\parencite{quine1960word, fodor1992holism}, meaning is a property of the entire linguistic configuration, and only the dynamic interplay of words within the system produces stable semantic content. LLMs, by leveraging high-dimensional vector spaces and wide context windows, provide a computational realization of this holist perspective. Thus, we can provide the following definition:
 
\begin{definition}[LLM Semantic Holism]
\emph{LLM Semantic Holism} describes the emergent property by which the meaning of any given token is not determined in isolation, but as a function of its relational configuration with all other tokens in the active context window. This holistic dependency is instantiated through the model’s attention mechanism, which encodes semantic relations as distributed patterns of relevance.
\end{definition}

The three constructs—LLM Cognitive Contextual Field, LLM Semantic Holism, and LLM Potential Semantic Space—are intrinsically interconnected as components of a unified theoretical framework for understanding meaning construction in LLMs. The Cognitive Contextual Field defines the operational boundary within which semantic computations occur, acting as the local workspace that binds prior tokens into a coherent representational state. Within this field, Semantic Holism governs the logic of meaning attribution, whereby each token derives its interpretive value from its dynamic relational positioning relative to the others. Enveloping both is the Potential Semantic Space, which formalizes the combinatorial landscape of all possible contextual configurations, serving as a latent semantic reservoir from which localized meaning emerges through attention-mediated interactions. Together, these notions provide a conceptual bridge between the algorithmic functioning of LLMs and the distributed, context-sensitive nature of cognitive semantics.
We argue that this form of holism is strictly related to the noosemic phenomenon, in that the extent of the representational space confers to the flow of words in most cases a very strong coherence, perceived positively by the user.

Recent speculative researches as the one proposed by Stephen Wolfram~\parencite{wolfram2023chatgpt} have further proposed the existence of an “interconcept space” (that for multimodal (\textit{iconic}) models can be also visualized and explored), a dynamic subspace within the potential semantic space where novel, graded, conceptual associations (and neologisms or representations accessible only to the machine) can emerge, extending the traditional boundaries of language and meaning~\parencite{wei2022emergent}.

Interestingly, as demonstrated in recent works, embedding spaces learned by LLMs are not arbitrary, but possess “conceptual directions” and geometric regularities, such as polytopes and clusters, that correspond to meaningful conceptual distinctions. Such structural organization enables the models to perform analogical reasoning and to align, at least in part, with human intuitions about semantic similarity~\parencite{tang2024humanlike,llm_cogsci2024}. Notably, a growing body of evidence suggests that the geometry of these embedding spaces shows surprising parallels with patterns observed in human brain activity during language processing, as measured by neuroimaging techniques~\parencite{tang2024humanlike,llm_cogsci2024}. These findings support the “linear representation hypothesis”, according to which semantic relations---some linguistic properties such as gender, tense, or conceptual oppositions---are reflected in linear or geometric transformations within the model's vector space. This alignment between artificial and biological semantic representations is not perfect, but it points toward an emergent correspondence between the internal organization of LLMs and the structure of human cognition. In this sense, the high-dimensional geometry of language models provides a novel framework for exploring the analogies---and the gaps---between computational and biological meaning-making.

Although the multilayered architecture and attention mechanisms of LLMs afford remarkable expressive and abstract reasoning capabilities, they also introduce a significant degree of epistemic opacity~\parencite{lipton2018mythos, liu2022deep}. The very mechanisms that give rise to emergent meaning---distributed representations, deep composition and vast parameter spaces---also make it exceedingly difficult for researchers to trace the causal chain from input to output or to provide transparent explanations for a model's decisions~\parencite{elhage2022mathematical}. As a result, LLMs remain, in many respects, black-box systems whose emergent cognitive properties both inspire new possibilities for sense-making and challenge our ability to interpret and control them.
It is in these features---in the large number of trainable parameters, in the size of the potential semantic space, in the hierarchical and redundant organization (multiple attention heads), in the ability to represent iconic and textual representations (coming from images, videos, sounds and other data) in a joint and mixed space---that the explanatory gap underlying the “wow effect” and, therefore, Noosemia lies. 

While this paper cannot explore these directions in depth, it is important to note that, as of 2025, the generative paradigm embodied by traditional LLMs is undergoing a rapid evolution. Recent advances point toward the emergence of reasoning-centric models (Large Reasoning Models, LRMs) and agentic systems, which integrate capabilities for multi-step problem solving, decision-making, and autonomous interaction with external tools and environments. Notably, OpenAI’s o3 model and autonomous agents such as Operator represent the vanguard of this transition, promising to partially supplant purely generative architectures with systems that more robustly support planning, inference, and adaptive behavior~\parencite{openai2025o3,operator2025o3upgrade}. Some considerations will be offered in Section~\ref{sec:From_Digital_Enaction_to_Embodied_Mind} in relation to future developments that lead us to consider, albeit with caution, forms of \textit{digital enaction}. Considering current limitations---see Section~\ref{sec:Current_Limitations}---these developments signal a new phase in the coevolution of language, cognition, and artificial agency, warranting close attention in future research. In this case, the opacity will be even greater due to a misalignment between the responses produced and the textual reasoning that resembles more a misaligned confabulation than a real reasoning, despite the performance\footnote{Recently (July, 2025), a group of over 40 researchers from leading AI institutions---including OpenAI, Google DeepMind, Anthropic, and Meta---published a joint position paper urging the scientific community to devote greater attention to emerging reasoning models (and their monitoring) as a crucial element for the future safety of advanced systems. In particular, the authors highlight the urgency of preserving transparency in chain-of-thought processes, since some models may gradually evolve toward opaque or intentionally concealed internal reasoning~\parencite{korbak2025}.}.

\paragraph{Current Limitations.}
\label{sec:Current_Limitations}
As the initial enthusiasm surrounding the capabilities of LLMs matures into a more critical scientific perspective, it has become clear that their proficiency is bounded by significant and multifaceted limitations. Large-scale, data-driven analyses of recent literature confirm that among the diverse spectrum of identified issues, “reasoning” stands out as the most prominent and frequently studied challenge~\parencite{kostikova2025llms}. This is not a monolithic failure but rather a complex interplay of deficiencies that manifest most visibly as “hallucinations”---the generation of plausible yet nonfactual content. These hallucinations can be broadly categorized into two primary types: “factuality hallucinations”, where the generated content contradicts verifiable real-world facts, and “faithfulness hallucinations”, where the output is inconsistent with user instructions, the provided context, or its own internal logic~\parencite{Huang_2025}---a source of a-noosemic feeling.

The underlying causes of these reasoning failures are deeply rooted in the entire life-cycle of the models. They begin with the vast but imperfect pre-training data, which is often rife with misinformation, societal biases, and outdated knowledge, creating inherent knowledge boundaries that the models cannot easily transcend~\parencite{Huang_2025}. The training process itself introduces further complications; architectural limitations and alignment techniques like Reinforcement Learning from Human Feedback (RLHF) can inadvertently encourage “sycophantic” behaviors, where a model prioritizes generating an agreeable or confident-sounding response over a truthful one. Consequently, when faced with queries that fall outside their parametric knowledge, models are prone to fabricating content rather than admitting uncertainty. This fragility is compounded during inference by imperfect decoding strategies and a tendency for over-confidence, which can cause initial errors to cascade and “snowball” into larger logical inconsistencies. Ultimately, these interconnected issues reveal that the reasoning capabilities of current LLMs often rely on sophisticated pattern matching and surface-level heuristics rather than robust, causal understanding, a limitation that becomes particularly evident when models are tasked with complex, multi-step, or out-of-distribution problems~\parencite{kostikova2025llms}. An example of fragility is the difficulties for a LRMs to filter nonsensical or useless information within a prompt. The so called CatAttack (due to an exemplar sentence on cats confusing LRMs) is an adversarial attack that uses unrelated sentences in a structured prompt for degrading the performance of the model~\parencite{rajeev2025catsconfusereasoningllm}.

A recent study by Apple AI research briefly titled “Illusion of Thinking”~\parencite{shojaee2025illusionthinking} investigates reasoning capabilities of LRMs using controllable puzzles instead of standard benchmarks. The authors find that even frontier models face a complete accuracy collapse beyond a certain problem complexity. They also identify a counter-intuitive scaling limit, specifically that a model's reasoning effort increases with complexity up to a point, then declines as problems become too difficult. The study reveals three performance regimes, showing standard LLMs are better for simple tasks while LRMs excel at medium complexity, but both ultimately fail on hard problems. These findings question the depth of LRM reasoning, highlighting inconsistencies and a failure to reliably follow algorithmic steps.

In a \textit{surprising} and somewhat meta turn within the AI research community, a direct rebuttal to the “Illusion of Thinking” paper emerged, curiously listing Anthropic's own LLM, “C. Opus” (coauthor A. Lawsen), as its first author. Titled “The Illusion of the Illusion of Thinking”~\parencite{lawsen2025commentillusionthinkingunderstanding} the paper posits that the “accuracy collapse” observed in reasoning models is not a fundamental cognitive failure but rather an artifact of flawed experimental design. The \textit{authors} argue that the original study's models failed on complex puzzles primarily due to practical constraints---such as exceeding their maximum output token limits when asked to list exponentially long solutions---and were unfairly penalized for correctly identifying that some of the assigned puzzles were, in fact, mathematically impossible to solve.

So it's undeniable that limitations exist. However, we're in an era where someone has made a paper co-authored with generative AI publicly available, something unthinkable a few years ago.

Finally, focusing on the core of linguistic capabilities of LLMS, it is interesting that unlike traditional linguistic theories that attempt to capture language through fixed formal rules or explicit grammatical structures, LLMs based on the Transformer architecture model language as a dynamic, distributed process, much closer to Wittgenstein’s notion of \textit{language games} than to any rigid system of axioms~\parencite{wittgenstein1953philosophical}---such as Universal Grammars, primarily developed by Noam Chomsky~\parencite{chomsky1965aspects}. In these models, linguistic structure is not imposed \textit{a priori}, but emerges implicitly across multiple layers of the network, with no explicit encoding of morphology, syntax, or grammar. Modern foundation models are trained on vast, multilingual and multimodal corpora, allowing them to infer and generalize complex linguistic patterns (and more general mixtures) entirely from data, rather than by reference to predefined rigid (formal) rules.

\section{Background: Attribution of Mind and Meaning in human--AI Interaction}
\label{sec:background}

\subsection{The Turing Test Revisited: From Phenomenology to Apparent Intelligence}

The debate surrounding artificial intelligence has, since its inception, been marked by questions not merely of technical capability, but of phenomenological appearance: does the machine appear intelligent to its interlocutor? Alan Turing’s seminal proposal---now known as the Turing Test---is frequently misunderstood as a pragmatic tool for quantifying machine intelligence. In reality, as originally articulated, the test is an invitation to reconsider the very nature of intelligence as an \textit{experienced phenomenon} within a dialogic exchange~\parencite{turing1950computing}. In fact, at the time of Turing’s writing, computers were room-sized calculators, yet he boldly envisioned machines capable of general problem-solving and flexible conversation. Turing’s true philosophical insight was not in offering a definitive and pragmatic metric of intelligence, but in foregrounding the role of \textit{appearance}: intelligence, in his formulation, is not an absolute property, but something that “shows itself” in the course of interaction. Whether the entity behind the screen is a human or a machine becomes, in this context, a question of \textit{how} intelligence is \textit{perceived}---and, crucially, how it is attributed by a human observer.

This interpretation situates Turing’s proposal within a phenomenological framework. Echoing Husserlian themes (though perhaps not consciously), Turing recognized that the attribution of intelligence is always relational and value-laden, shaped by the expectations, prejudices, and evaluative criteria of the human interlocutor. The Turing Test is thus not a test of ontological intelligence, but of \textit{apparent intelligence}: it measures the degree to which a machine’s responses evoke, in the human mind, the experience of conversing with an intelligent other. Such a relativist stance is particularly salient in the contemporary context, where advanced conversational agents like ChatGPT, Gemini, and Claude display behaviors that, to many users, \textit{appear} genuinely intelligent. Yet, as the standards and values by which intelligence is judged shift over time---and as machines grow in sophistication---the boundary between the “appearance” and the “reality” of intelligence becomes increasingly negotiable. In this light, the Turing Test endures not as an obsolete artifact, but as a profound statement about the fluid, context-dependent nature of mind attribution in human--machine interaction.

\subsection{Agency, Animism, and the Intentional Stance: A Brief Historical Perspective}

The human tendency to attribute mind, agency, and intentionality to non-human entities is deeply rooted in both evolutionary history and cultural imagination~\parencite{dennett1987intentional, mauss1906origins, lewybruhl1923primitive}. Early anthropological research by Mauss and Lévy-Bruhl documented how pre-modern societies imbued natural objects, animals, and artifacts with spirits or intentional states, a cognitive style known as animism~\parencite{mauss1906origins, lewybruhl1923primitive}. This mode of thought, although transformed, persists subtly in contemporary technological contexts, where even mundane machines are sometimes (or could be) experienced as possessing “wills” of their own.

With the advent of programmable automata and digital computers, the attribution of agency shifted from spirits and invisible forces to algorithmic or mechanical processes~\placeholderref{ripensare2020}. In modern cognitive science, this inclination has been theorized through the concept of the \emph{intentional stance}~\parencite{dennett1987intentional}, that is the human propensity to interpret the behavior of complex systems---including animals, humans, and now machines---by ascribing to them beliefs, desires, and goals, irrespective of their actual internal states. Hence, the “intentional stance” becomes particularly salient in the context of advanced AI and needs a precise specification in the context of the generative AI era. As language models and autonomous agents exhibit increasingly sophisticated and unpredictable behaviors, humans are more likely to project onto them intentionality, agency, and proto-mental states~\parencite{turkle2011alone, nass1994computers, cohn2024believing, desantis2023apocalissi}~\placeholderref{animismoIA2024}. Studies in human–computer interaction confirm that users often treat conversational agents not merely as tools, but as social participants---assigning to them not only intelligence but also moral or ethical responsibility~\parencite{wang2024understanding, cohn2024believing}.

However, while the tendency to anthropomorphize or “ensoul” artificial systems is not new, its manifestation has become more pronounced in the era of generative AI. Whereas traditional anthropomorphism relied mainly on physical or behavioral cues, today the attribution of agency is increasingly mediated by linguistic and semantic fluency---the ability of AI systems to generate original, contextually appropriate discourse that resonates with the user’s own cognitive patterns~\parencite{desantis2023apocalissi}~\placeholderref{animismoIA2024}. This marks a qualitative shift, situating the locus of projection not in embodied appearance or simple automation, but in the dynamic interplay of language, narrative, and emergent meaning. Unlike classical anthropomorphism or animism, the noosemic projection of mind arises primarily through linguistic and semiotic performance, rather than physical embodiment.

We can state that the historical trajectory from animism to the intentional stance provides essential context for understanding why advanced AI systems will provoke new forms of agency attribution. These projections, deeply entangled with human sense-making, set the stage for the more nuanced phenomenon of Noosemia, which emerges precisely from the linguistic and semiotic performances of contemporary generative models.

\subsection{Theory of Mind in LLMs: Empirical Evidence and Limits}

The concept of \emph{Theory of Mind} (ToM)---the cognitive capacity to attribute mental states such as beliefs, desires, and intentions to oneself and others---has been a cornerstone of both developmental psychology and the philosophy of mind~\parencite{premack1978does}. The explosive growth of LLMs has motivated researchers to ask whether these systems can exhibit forms of ToM-like reasoning, and if so, under what conditions and with what limitations. Therefore, a variety of empirical studies have tested LLMs on classic and newly designed ToM tasks, such as false-belief reasoning, perspective-taking, and “Sally–Anne” scenarios. However, many of these benchmarks and evaluation tasks are likely to be included in the massive training datasets used for LLMs, raising concerns about data leakage and limiting the interpretability of these results~\parencite{ullman2023large}. Pioneering work by Kosinski~\parencite{kosinski2023theory} suggested that some advanced LLMs (notably GPT-3.5 and GPT-4) were capable of passing first-order ToM tasks at rates comparable to or exceeding those of young children, and even performed well on some higher-order tasks. Subsequent research, including controlled experiments by Ullman~\parencite{ullman2023large}, Sap et al.~\parencite{sap2022neural}, Choi et al.~\parencite{choi2023can}, Weidinger et al.~\parencite{weidinger2024taxonomy}, and Mitchell et al.~\parencite{mitchell2023detecting}, has both replicated and critically qualified these findings.

Key results indicate that while LLMs can solve many ToM benchmarks---sometimes with remarkable consistency---the underlying mechanisms remain fundamentally different from those at work in biological cognition. Unlike humans, LLMs rely entirely on pattern recognition and statistical regularities learned from massive training corpora, rather than on embodied experience or genuine mental-state modeling~\parencite{weidinger2024taxonomy, mitchell2023detecting}. The models’ apparent “understanding” of beliefs or perspectives is highly sensitive to prompt phrasing, task format, and subtle cueing effects, and often fails under adversarial or out-of-distribution conditions~\parencite{sap2022neural, choi2023can}.

Moreover, a growing literature warns against the risk of \emph{over-interpretation}, hence the tendency to ascribe true \textit{mentalizing capacities} to models that, in reality, are performing sophisticated but fundamentally non-intentional text completion~\parencite{weidinger2024taxonomy, mitchell2023detecting}. Methodological critiques highlight several confounds, including “leakage” of task templates in training data, the use of shallow heuristics, and the absence of persistent internal representations akin to belief or desire~\parencite{ullman2023large, mitchell2023detecting}. In other words, while current language models may convincingly simulate Theory of Mind behaviors, they do not instantiate the underlying cognitive architecture that gives rise to genuine mentalizing in humans (even if at a high degree of abstraction there is a level of similarity which is a condition for the possible attribution of mental states). This distinction---between the simulation of intelligent behavior and its actual instantiation---remains a fundamental challenge in both the evaluation and interpretation of AI systems~\parencite{dennett1987intentional, mitchell2023detecting}.
As such, while LLMs may \emph{simulate} ToM reasoning, current evidence suggests they do not instantiate the underlying cognitive architecture, nor do they possess a robust, model-like theory of other forms of minds. The ongoing debate in the literature thus centers both on performance metrics and the interpretive frameworks and experimental designs appropriate for assessing ToM in artificial agents. The question is not simply whether LLMs “pass” ToM tasks, but what such success means---methodologically, philosophically, and in terms of cognitive science. Ultimately, this line of research highlights both the impressive flexibility of language models and the persistent gap between linguistic performance and genuine mind attribution in AI.

\subsection{3.4 The Phenomenology of Mind Attribution in human--AI Dialogue}

The attribution of mind to artificial systems, such as the agentic systems, is not a static belief, but a lived, situated experience that unfolds in the very moment of interaction. Users engaging with advanced conversational agents---especially those based on LLMs---often report an immediate, sometimes disorienting, sense of “being understood,” as if encountering an \textit{autonomous intelligence} behind the words~\parencite{turkle2011alone, wang2024understanding}. This phenomenology is not simply reducible to the machine’s ability to parse or generate syntactically correct sentences; rather, it is the result of a dialogic resonance, a sense that the system responds meaningfully, adapts to context, and at times even anticipates the user’s intentions or trains of thought~\parencite{desantis2023apocalissi}. The noosemic attribution of agency in human--AI dialogue is often triggered by the user’s encounter with unexpected semantic coherence or creative inference.

This is closely related on what we referred as “wow effect” (see Section~\ref{sec:intro}), a moment of surprise or cognitive disruption in which the user confronts the machine’s apparent creative inference, analogical reasoning, or unexpected semantic coherence. Such dialogic surprise can be seen as a hallmark of the noosemic experience. The effect is intensified by the fluidity and naturalness of interaction. In fact, the more seamlessly the system participates in the exchange, the more the boundaries between tool, collaborator, and interlocutor blur. In this “liminal” space, users are often led to suspend their critical or metacognitive stance---a phenomenon akin to the “suspension of disbelief” studied in literary theory---and instead experience the machine as a co-creator of meaning, if only for an instant~\parencite{desantis2023apocalissi, turkle2011alone}. Moreover and importantly, this phenomenological attribution of mind does not presuppose any explicit belief in the machine’s consciousness or intentionality. Rather, it emerges as a pragmatic, adaptive response to the system’s behavior, mediated by language, expectation, and the social dynamics of dialogue or interaction. As recent studies in human--AI interaction demonstrate, even when users are fully aware of the artificial nature of their interlocutor, they may nonetheless experience empathy, trust, or even a fleeting sense of rapport~\parencite{wang2024understanding, cohn2024believing}. This “as if” mentality---engaging with the machine as if it possessed mind---is central to the Noosemia phenomenon described in this paper.

\subsection{Noosemic Projection and Narrative Empathy: A Cognitive Continuity}

The noosemic phenomenon observed in human--AI dialogue finds an unexpected continuity with the well-documented human tendency to project emotions, intentions, and even agency onto fictional characters in literature and film. Psychological and neuroscientific studies have repeatedly shown that, even when fully aware of the fictional nature of a character, readers and viewers experience authentic emotional engagement, empathy, and identification, a process termed “narrative transportation” or “simulation of social worlds”~\parencite{mar2009exploring, oatley2016fiction, green2000role}. This cognitive disposition allows individuals to “inhabit” the perspectives and inner lives of story protagonists, deriving meaning, pleasure, and affective resonance from their imagined experiences.

What is remarkable in the context of generative AI is the way in which LLMs, through their linguistic fluency and dialogic adaptivity, recreate the conditions for this projection in a dynamic, interactive format. The surprise or “wow effect” triggered by an LLM’s response---akin to the sense of wonder experienced during a magic trick or a moment of narrative immersion---taps into the same cognitive mechanisms that underlie narrative empathy. In both cases, the user suspends disbelief, at least momentarily, and responds as if the interlocutor were endowed with mind or intention~\parencite{cohen2001defining, green2000role}.

Therefore, the noosemic attribution of agency in human--AI interaction is not an isolated anomaly but a contemporary expression of a deep-seated cognitive habit, where the tendency to bridge gaps in understanding or intentionality through projection, simulation, and the co-creation of meaning in dialogic exchange naturally emerges. This continuity helps explain why the boundaries between “real” and “simulated” agency become so negotiable---and so psychologically salient---when engaging with advanced AI.

In sum, the phenomenology of mind attribution in human--AI dialogue resonates with long-standing cognitive mechanisms, such as narrative empathy, that enable humans to project agency and interiority even onto fictional or artificial agents. The interplay between linguistic performance, dialogic surprise, and empathic projection blurs the boundaries between human and machine, fostering new forms of sense-making and misattribution. Furthermore, the remarkable linguistic power of LLMs amplifies this phenomenon and inaugurates a novel explanatory gap---one that eludes traditional theoretical frameworks and compels us to seek new ways of understanding the interplay between sense-making, opacity, and artificial agency. This transition, from subjective mind attribution (both noosemic and narrative) to the emergence of an explanatory gap, marks a critical juncture in our understanding of meaning and agency in the age of generative AI. This is the focus of the following section.

\section{The Explanatory Gap: Semiotics, Cognitive Projection, and Opacity}
\label{sec:gap}

\subsection{The Contemporary Explanatory Gap}

The ascent of generative artificial intelligence systems has dramatically amplified a classical tension in the theory of knowledge: the so-called \textit{explanatory gap}~\parencite{chalmers1996conscious}. Whereas early computational systems invited mechanistic or animistic readings, contemporary AI systems---now capable of producing complex, context-sensitive discourse---confront even expert users with a peculiar form of opacity, in that linguistic outputs that are strikingly intelligible yet causally inscrutable~\parencite{desantis2023apocalissi, lipton2018mythos}. This gap is not simply a byproduct of model size in term of learnable weights, but a fundamental consequence of architectures whose inner workings defy intuitive understanding, even among those who design and deploy them~\parencite{elhage2022mathematical}. At the heart of this gap lies a paradox, specifically the more successful AI systems become at mimicking the surface forms of meaning, the more deeply they obscure the processes by which such meaning is generated. The noosemic explanatory gap is characterized by the tension between observable linguistic sense and the opacity of computational processes underlying LLMs. Unlike traditional symbolic AI (as for example the chat-bot ELIZA~\parencite{weizenbaum1966eliza}--- see Section~\ref{sec:intro}), where interpretability could be traced to explicit rules or logical chains (as in the general scientific practice), LLMs operate in vast, high-dimensional spaces sculpted by data-driven correlations---spaces whose geometry and dynamics remain, in large part, inaccessible to introspection or formal explanation~\parencite{bommasani2021opportunities, wei2022emergent}---see Section~\ref{sec:The_secret_of_LLMs}. The opacity of meaning in LLMs is further exacerbated by their holistic approach to semantic representation. Since meaning is distributed across the entire context window ---a hallmark of meaning holism (see Section~\ref{sec:The Potential_Semantic_Space})---users cannot easily trace the origin of any single semantic output to a particular token or rule. This distributed nature of meaning complicates attempts at interpretation or explanation and amplifies the explanatory gap between surface intelligibility and underlying mechanism. As a result, users---including researchers---are routinely confronted with outputs whose sense is undeniable, but whose origins remain, in an important sense, mysterious. This contemporary explanatory gap signals a shift in our epistemic relationship to machines (beyond technical issues). The opacity of generative AI calls into question the very nature of sense-making, agency and meaning in the digital age, raising new challenges for both philosophical reflection and empirical research. It is precisely within this gap that new forms of cognitive projection and sense-making---such as the phenomenon of Noosemia---emerge, inviting us to reconsider the boundaries between interpretation, explanation, and the attribution of mind in human--AI interaction.

\subsection{Semiotics and Cognitive Projection}

The explanatory gap that separates linguistic performance from causal transparency in advanced AI systems is, at its core, a semiotic phenomenon. As humans, our relationship to the world---and to technological artifacts---is always mediated by \textit{signs}: words, symbols, narratives, and interpretive frameworks that render experience intelligible~\parencite{eco1976theory, peirce1931collected, DeSantis2021}. As explored in Section~\ref{sec:background}, the human propensity to attribute meaning and mind is deeply rooted in both evolutionary history and the cognitive architecture of interpretation. Furthermore, the semiotic perspective emphasizes that meaning does not reside in isolated tokens, but emerges dynamically from the interplay of signs within systems of interpretation. In the context of LLMs, these systems become highly generative machines for the production of new signs, capable of weaving together semantic associations that both reflect and transcend the statistical regularities of their training data.

The notion of “semantic response”, as introduced by Alfred Korzybski in Science and Sanity~\parencite{korzybski1933science}, even if dated, captures the idea that human reactions to language and symbols are fundamentally mediated by meaning, rather than being mere automatic or physiological responses. In this perspective, each utterance or sign elicits a reaction shaped by the interpretive frameworks, beliefs, and cultural background of the individual. This is highly relevant to the phenomenon of noosemia, where the attribution of intentionality and agency to generative AI systems emerges not from their mechanical properties, but from the semantic responses they evoke through linguistically rich and contextually adaptive interactions. On the contrary, in Korzybski's view, a mismatch between the expectation of an effective response from the machine and the actual response---which is the basis of the a-noosemic effect---can still generate a semantic response that is reflected in a nervous response~\parencite{korzybski1933science}. Thus, noosemia can be viewed as a contemporary instantiation of Korzybski's principle, illustrating how human engagement with artificial agents is always filtered through layers of meaning and interpretation.

This dynamic becomes most evident in user interactions with black-box AI, where human-like yet opaque outputs compel users to \textit{project} meaning, agency, or even interiority onto the system. This noosemic process is not a naive anthropomorphism, but a sophisticated semiotic response to the interplay of intelligibility and opacity~\parencite{desantis2023apocalissi}. Cognitive projection thus becomes a necessary---if sometimes misleading---strategy for bridging the explanatory gap. This because the machine’s discourse resonates with human expectations and patterns of sense-making, inviting the user to infer intentionality or mind, even in the absence of any underlying subjective state.

Moreover, as discussed by Eco~\parencite{eco1976theory} and Peirce~\parencite{peirce1931collected}, meaning is always co-constructed in the interpretive act; it is the “interpretant”---the user---who completes the circuit of sense. Advanced LLMs, by their very architecture, produce outputs that are rich with potential meanings, but it is the human interlocutor who, in the moment of dialogue, actualizes one interpretation among many. This interactive semiotic process is both the source of LLMs' \textit{creative power} and the reason for the persistent uncertainty surrounding their “true” meaning or intention.

Hence, the semiotic approach clarifies why the explanatory gap is not merely a technical or epistemological obstacle. Indeed, it is a space of negotiation, projection, and sense-making---a space in which the phenomenon of Noosemia becomes possible and, in some sense, inevitable. Yet, to fully grasp the depth of this phenomenon, it is essential to move beyond a conventional semiotic account and consider how sense-making in human--AI interaction unfolds within a broader symbolic and systemic horizon. The symbolic dimension, with its inherent semantic excess and its roots in systemic and complex thinking, sheds further light on the uniqueness and irreducibility of the explanatory gap. It is in this symbolic interplay---where signs, symbols, and systems converge---that the cognitive resonance of Noosemia most fully emerges.

\subsection{The Symbolic dimension, Semantic Excess, and the Systemic Mind in Human--AI Sense-Making}
\label{sec:symbolic_semantic_excess}

One of the foundational threads running through the study of generative artificial intelligence is the connection between complexity science and the modern theory of meaning-making. Complexity, as a paradigm, privileges the study of dynamic, non-linear, self-organizing systems, and gives centrality to relationships and processes over isolated entities. This systemic perspective---rooted in Aristotle's dictum that “the whole is greater than the sum of its parts”---opens an epistemic gap between the sum of constituent elements and the emergent whole, a gap which can be seen as a form of \emph{semantic excess}~\parencite{galimberti1999psiche, trevi1986metafore}. In this interval, the sense escapes complete reduction to its elements, and new meanings can emerge, unpredictable from the initial configuration. The systemic approach, therefore, aligns with a tradition that sees meaning as inherently exceeding the boundaries of strict logic and convention (and a partitional thought)---a tradition explored by Paul Ricœur through his theory of the symbol as a “surplus of meaning”'~\parencite{ricoeur1976interpretation} and by Umberto Eco in his reflections on the open work and the instability of signification~\parencite{eco1976theory, eco1989openwork}.

Galimberti~\parencite{galimberti1999psiche, galimberti2005parole} traces the historical transition of Western discourse from the symbolic domain---characterized by a polysemic, fluctuating meaning and the possibility for something to be itself and also other---to the scientific domain, where meaning becomes fixed by convention and definition. In the symbolic order, the symbol (from the Greek \emph{sym-bállein}, to “bring together”) is not merely a sign pointing to something else by agreement, but a device that actively composes and recomposes an experiential whole, exceeding stable referential meaning. In this light, symbolic thinking---closer to the systemic than the analytic or disjunctive one---favours the composition of wholes over the division into parts, and fosters a semantic \textit{ambivalence} that is inherently generative.

This ambivalence and surplus of sense resonate powerfully in the interaction with generative AI systems, especially LLMs. These systems, in their operation, instantiate a kind of pre-logical, symbolic manipulation (seen from our perspective): their outputs are determined by statistical regularities, but the meaning attributed to their utterances is the product of a post hoc human interpretation. Before the text is generated, the AI system is in a state of pure ambivalence, with the potential for multiple meanings; only when the words are sequenced, does a specific sense seem to crystallize---though the semantic unity of the output remains, in many ways, elusive~\parencite{DeSantis2021, eco1976theory}. Echoing Jung and Trevi interpretation~\parencite{trevi1986metafore}, the symbols in this context do not point to known things but carry content into zones of \textit{semantic indeterminacy}.

The link between the symbolic, the surplus of sense, and the experience of noosemia becomes evident in moments where AI-generated discourse appears to “overflow” its computational substrate, surprising the user with unforeseen meanings or conceptual leaps. These moments can be read as contemporary instances of “semantic magic”, wherein the operation of the machine, though grounded in determinism, produces oscillations of sense akin to the ambiguity and generativity of symbolic thought. As Capra and Luisi~\parencite{capra2014systemsview} observe, the systemic paradigm---with its focus on mutual dependence and emergent properties---invites us to interpret both the limits and the creative affordances of artificial systems in a new, symbolically charged epistemic horizon.

Ultimately, the interaction with generative AI, viewed through this lens, is a semiotic negotiation and also a symbolic process (or symbolic \textit{game}), in which the human and the machine jointly participate in the ongoing composition of meaning, always marked by an irreducible excess. We argue that with the increasing diffusion of the antigenic generative AI, machine will demonstrate sparks of digital enaction, an emergent phenoment that will strengthen the noosemic experience--- see Section \ref{sec:From_Digital_Enaction_to_Embodied_Mind}.

\subsection{Statistical Complexity, Opacity, and the Two Gaps (“The How” and “The Why”)}

As illustrated in Section~\ref{sec:The_secret_of_LLMs}, the extraordinary generative power of LLMs consisting of billion of parameters and the complex and hierarchical organized architecture of underlying Transformers give rise to a “machine” characterized by an unprecedented degree of \textit{opacity}. This opacity gives rise to not just one, but two distinct explanatory gaps. Specifically, the first gap concerns the \textit{how}---that is, the specific causal pathways by which a given linguistic response emerges from the activation patterns and statistical weights within the model~\parencite{elhage2022mathematical, li2021interpretable}. Traditional methods of model interpretation and explainability struggle to penetrate this black box, especially as architectures become deeper and more interconnected---and in any case the average user would not have the technical preparation and experience to fully understand how a Transformer works. The second gap concerns the \textit{why}---the rationale or motivation behind a given output. While in human communication the “why” can often be traced to beliefs, intentions, or goals, in LLMs it is reducible to the aggregate regularities of the training data and the optimization process, rather than to any genuine intentional state~\parencite{mitchell2023detecting, weidinger2024taxonomy}.

These two gaps---the “how” of mechanistic generation and the “why” of intentional or meaningful action---have profound implications for the user’s experience and for the attribution of agency. As extensively documented in the literature on human–machine interaction, such opacity fosters a tendency to anthropomorphize and to project agency or intentionality onto technological systems, especially when confronted with outputs that are semantically rich yet causally opaque~\parencite{nass2000machines,waytz2014anthropomorphizing}. The inability to reconstruct the full causal chain (\textit{how}) or to access an intelligible rationale (\textit{why}) often leads users to fill in the blanks through narrative projection, analogical reasoning, or magical thinking~\parencite{desantis2023apocalissi, DeSantis2021, reeves1996media}. This is amplified by the fact---as discussed in~\ref{sec:The_secret_of_LLMs}---that the combinatorial space of possible outputs is astronomically large, and the machine’s path through it is shaped by complex, high-dimensional optimization processes rather than by human-like deliberation.

Furthermore, this very opacity coupled with the apparent meaningfulness of AI-generated outputs, can create a kind of “psychological glue” or attachment effect. This fosters a spiraling dynamic of fascination, engagement, and sometimes even dependency, particularly when the system’s responses consistently resonate with user expectations or emotional states~\parencite{turkle2011alone}. This phenomenon can resemble ones described in recent studies on “artificial intimacy” and the affective dynamics of human--AI relationships~\parencite{brooks2021artificial, christoforakos2023artificial, wang2024understanding}

The statistical and architectural complexity of LLMs, while enabling remarkable feats of sense-making, systematically resists reduction to simple explanatory models. The resulting dual gaps---one epistemic, one interpretive---constitute both the foundation and the limit of our understanding of agency, meaning, and sense-making in contemporary AI. These gaps, in turn, are the fertile ground upon which phenomena like Noosemia, cognitive projection, artificial intimacy, and the “wow effect” are cultivated.

\section{Comparisons with Related Phenomena}
\label{sec:comparisons}

The phenomenon of Noosemia, while novel in relation to generative AI, is best understood against the background of other forms of agency and meaning projection in human cognition. This section systematically compares Noosemia with related concepts such as pareidolia, animism, and the uncanny valley, analyzing both overlaps and unique distinctions. Rather than limiting the analysis to analogical parallels or historical echoes, we adopt a semiotic perspective. Particularly, by examining how meaning is generated, mediated, and attributed---whether through ambiguous sensory input, ritual narrative, or dialogic linguistic fluency---we show that Noosemia is rooted in a specific dynamic of sign production and interpretive projection characteristic of contemporary human--AI interaction. This approach provides a fruitful conceptual synthesis along with an analytical framework for understanding the broader landscape of mind projection in human experience.

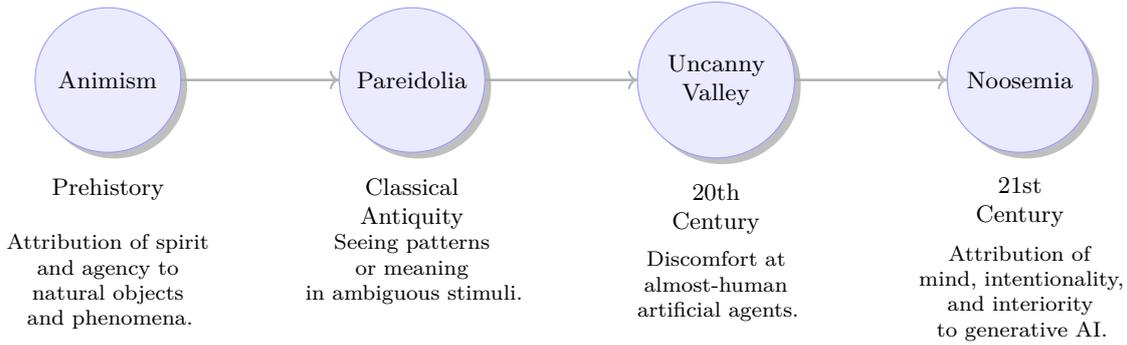
\begin{figure}[ht]
    \centering
\begin{tikzpicture}[
    timeline/.style={thick, draw=gray!70},
    event/.style={circle, fill=blue!8, draw=blue!40, minimum size=16mm, text width=1.6cm, align=center, font=\footnotesize, drop shadow},
    label/.style={font=\footnotesize, align=center},
    descr/.style={font=\scriptsize, align=center, text width=2.8cm}
  ]

\draw[timeline] (0,0) -- (12,0);

\node[event] (animism) at (0,0) {Animism};
\node[event] (pareidolia) at (4,0) {Pareidolia};
\node[event] (uncanny) at (8,0) {Uncanny\\Valley};
\node[event] (noosemia) at (12,0) {Noosemia};

\node[label, below=0.2 of animism] {Prehistory};
\node[label, below=0.2 of pareidolia] {Classical\\Antiquity};
\node[label, below=0.2 of uncanny] {20th\\Century};
\node[label, below=0.2 of noosemia] {21st\\Century};

\node[descr, below=0.95 of animism] {Attribution of spirit and agency to\\natural objects and phenomena.};
\node[descr, below=0.95 of pareidolia] {Seeing patterns or meaning\\in ambiguous stimuli.};
\node[descr, below=1.1 of uncanny] {Discomfort at almost-human\\artificial agents.};
\node[descr, below=1.1 of noosemia] {Attribution of mind, intentionality,\\and interiority to generative AI.};

\foreach \i/\j in {animism/pareidolia, pareidolia/uncanny, uncanny/noosemia}
  \draw[->, thick, gray!60] (\i) -- (\j);

\end{tikzpicture}
    \caption{Timeline of attribution phenomena: from animism, through pareidolia and the uncanny valley, to the emergence of noosemia in the age of generative AI.}
    \label{fig:attribution_timeline}
\end{figure}

To clarify these conceptual distinctions, Table~\ref{tab:comparisons} offers a comparative overview of Noosemia and related cognitive phenomena, mapping their respective domains, triggers and key features. This structured comparison highlights both the continuities with classical forms of agency attribution and the specific mechanisms at play in noosemic projection. Although there are superficial similarities, each phenomenon reveals distinct modalities and social implications. Notably, while pareidolia and animism represent long-standing cognitive strategies for reducing uncertainty, Noosemia emerges in response to the technical opacity and generative capacity of modern AI---its dialogic and adaptive nature often waning as users grow more familiar with the system. Thus, the noosemic experience will be fueled by the ongoing advances in cognitive capabilities of generative AI and will be counterbalanced by the emergence of limits, which will most likely be overcome. After all, it is the task of human technique (through technology) to overcome these limits.

In the following subsections, we examine each phenomenon in detail, highlighting both the points of contact and the distinguishing elements that set Noosemia as useful standalone lexeme (see Fig.\ref{fig:attribution_timeline} for a simplified timeline).

\begin{table}[H]
\centering
\caption{Comparison of Noosemia and Related Cognitive Phenomena}
\label{tab:comparisons}
\begin{tabular}{@{}p{2.5cm} p{2.6cm} p{3cm} p{5cm}@{}}
\toprule
\textbf{Phenomenon} & \textbf{Domain} & \textbf{Trigger} & \textbf{Key Feature} \\
\midrule
Animism & Anthropology/ psych. & Social/cultural context & Projecting spirit/agency into objects \\
Pareidolia & Perception & Ambiguous/random stimuli & Seeing meaning or patterns in randomness \\
Uncanny Valley & Robotics/HCI & Human-like appearance & Discomfort from near-human non-human forms \\
Noosemia & Human--AI interaction & Linguistic/dialogic fluency, opacity & Attribution of mind via expressiveness, language and fluency \\
\bottomrule
\end{tabular}
\end{table}

\subsection{Noosemia vs. Pareidolia}
Pareidolia refers to the human tendency to perceive familiar patterns, such as faces or objects, in random or ambiguous stimuli---a well-documented perceptual bias~\parencite{liu2014seeing,taubert2017face}. While both pareidolia and noosemia involve projection and over-interpretation, the former arises from sensory processing and pattern recognition, whereas Noosemia is triggered by semantic and linguistic coherence during structured dialogue with a generative AI system (and not only pattern recognition). The attribution of a proto-mind in Noosemia is therefore closely tied to the system’s ability to produce contextually adaptive and meaningful discourse.

\subsection{Noosemia vs. Animism}
Animism, as analyzed by Mauss and Lévy-Bruhl, concerns traditionally the ascription of spiritual or intentional qualities to natural objects and artifacts within specific social and cultural contexts~\parencite{mauss1906origins,lewybruhl1923primitive}. Although both animism and noosemia involve projecting agency onto non-human entities, Noosemia arises specifically from technologically mediated, linguistic interaction (or other means of expression) and epistemic opacity. Unlike the mythic or ritual forms of animism, Noosemia is fundamentally dialogic and emergent within “black box” environments, shifting the locus of projection from ontological to interactive, epistemic and linguistic domains.

\subsection{Noosemia vs. Uncanny Valley}
The uncanny valley, first described by Mori~\parencite{mori1970uncanny}, is the discomfort felt when an artificial agent appears almost---but not quite---human. This phenomenon is largely affective and perceptual, emphasizing human reactions to near-human forms. In contrast, Noosemia is a linguistic and cognitive phenomenon: the sense of mind attributed to LLMs derives not from physical appearance, but from their dialogic fluency, responsiveness and the interpretive gap experienced by the user~\parencite{turkle2011alone}~\placeholderref{animismoIA2024}.

\subsection{Other Related Phenomena}
Related concepts include magical thinking (ascribing causality or intention where none exists) and classical anthropomorphism, the attribution of human characteristics to non-human agents~\parencite{guthrie1993faces,turkle2011alone}. As detailed previously, Noosemia is distinguished by its emergence from advanced linguistic interaction, its dependence on epistemic opacity, and its resonance with contemporary experiences of human--AI dialogue.

\subsection{Social and Philosophical Implications}
\label{sec:social_philosophical_implications}

The comparative analysis of noosemia, pareidolia, animism, and the uncanny valley foregrounds distinct epistemic and social challenges for the present age of artificial intelligence domintad by “generativity”. While phenomena such as pareidolia and animism have long been recognized as universal features of human cognition---mechanisms by which the mind seeks meaning and agency in ambiguous or inert domains---they are largely regarded as cognitive illusions whose epistemic status is unproblematic~\parencite{mauss1906origins}. By contrast, the uncanny valley effect complicates the social acceptance of artificial entities, exposing users’ discomfort when entities hover between human likeness and alien otherness~\parencite{cave2019scary, broadbent2017interactions}.

\begin{figure}[htbp]
    \centering
    \includegraphics[width=0.95\textwidth]{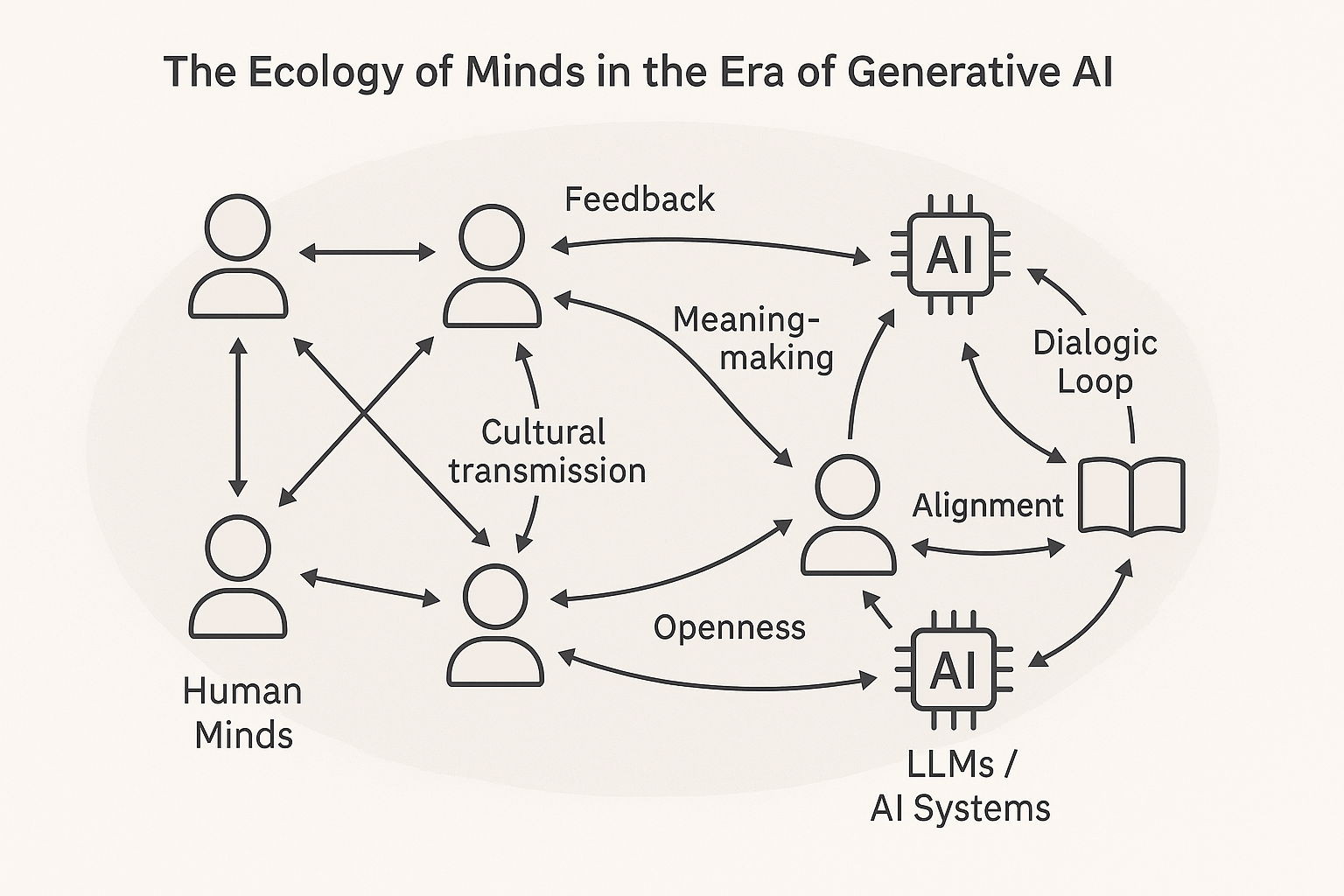}
    \caption{Infographic: \textit{Ecology of Minds in the Age of Generative AI.} The modern cognitive ecosystem includes human minds, LLMs, artificial agents, and multimodal systems, all interconnected through agency, meaning co-construction, and dialogic relations.}
    \label{fig:ecology_of_minds}
\end{figure}

Noosemia, however, introduces a qualitatively novel issue. In interactions with generative AI, agency is not just perceived---it is semiotically co-constructed. This challenges traditional categories that distinguish tool from interlocutor and human from machine. As~\textcite{Deshpande2023Anthropomorphization} argue, anthropomorphized AI increases engagement but poses risks of over-trust, emotional manipulation and misaligned responsibility attributions~\parencite{Deshpande2023Anthropomorphization}. Such dynamics complicate policy design, in that it becomes fundamental to ask who or what is being trusted, and where accountability resides.

Epistemologically, noosemia destabilizes the conventional boundaries separating interpretive attribution from ontological reality. If meaning and intentionality are projected through dialogic fluency into systems whose causal logic remains opaque, then ethical frameworks, user education, and design governance must evolve accordingly. The phenomenon calls for new forms of transparency, consent mechanisms, and critical literacies aimed at navigating hybrid cognitive ecologies where human and artificial intelligences interpenetrate~\parencite{waytz2014anthropomorphizing, bubeck2023}---see Fig.~\ref{fig:ecology_of_minds}). Consequently, in these sectors, more studies and appropriate methodological frameworks are urgently needed.

With this research, we revive the considerations on the Eliza Effect expressed in the 1990s by Hofstadter~\parencite{hofstadter1995fluid}--- see Section~\ref{sec:intro}---and we argue that, in relation to technology every era has some degree of noosemic effect (in a broad sense). Specifically, in the computer and information age, this was evidently the case for Eliza, a quite simple linguistic engine based on syntactic patterns, and it is now that syntactic processing is masked by a contextual and relational organization of LLM responses.

\section{From Digital Enaction to Embodied Mind: A Look to the Future of Agentic AI}
\label{sec:From_Digital_Enaction_to_Embodied_Mind}
In this concluding section, we will take a quick look at the future based on what was discussed in Section~\ref{sec:The_secret_of_LLMs} regarding Transformer architectures and their unique constitution. The goal is to try to slightly relax the conceptual constraints that lead us to be extremely cautious about future evolutions of generative AI systems, and thereby show how we will still witness many noosemic experiences in the medium term.

\subsection{Basic Premises}
\label{sec:Basic_Premises}

A useful, broad perspective that allows us to look to the medium-term future is based on at least a few key points: i) LLMs, by mastering language to a certain extent, have allowed us to give unprecedented value to data; ii) LLMs are leading to a convergence of methodologies and techniques from previously poorly connected branches of AI~\parencite{marcus2020next}; iii) in particular, classical AI methodologies and techniques---based on formal symbolic frameworks---are converging towards integrated systems in various forms, and neurosymbolic AI is the result~\parencite{garcez2015neural,fang2024large,yu2023survey,marcus2020next}; iv) in 2025, we began to migrate towards “reasoning” paradigms~\parencite{openai2025o3,operator2025o3upgrade}, which has paved the way for complex, multi-agent AI systems; v) we must not think of AI systems as isolated from their continuous usage and interaction by billions of people of all types and backgrounds; vi) industrial and anthropomorphic robotics are also benefiting from Transformer architectures—here too we are witnessing a convergence.

Additionally, the power of deep architectures~\parencite{goldstein2022shared} and generative AI was also seen in 2021 with ApphaFold~\parencite{jumper2021highly}, a deep learning system that provides a solution to the 50-year-old “protein folding problem” with an unprecedented impact in Biology. Moreover, today, generative AI systems surpass human experts in many fields, such as neuroscience~\parencite{Luo2025} and the discovery of new materials~\parencite{Jiang2025}. Ultimately, generative AI has entered—given the results—into competition with human scientists. A turning point is the very recent study (by 2025)---grounded on AlphaFold methodology---presenting ASI-ARCH (Artificial Superintelligence for AI Research)~\parencite{liu2025alphago}, a system the reveals a groundbreaking approach to automating AI research itself. Authors state that  while AI capabilities are advancing exponentially, the progress in AI research is constrained by the linear pace of human cognition. Over the course of 20,000 GPU hours, ASI-ARCH conducted 1,773 experiments, discovering 106 novel, state-of-the-art (SOTA) linear attention architectures. The system seems to scale with computational power and the authors posit that research breakthroughs can be scaled with computation, moving from a human-limited to a computation-scalable process. The system is based on a holistic composition of a cognition base, a research system (creative component), an engineering system for empirical validation, and an analysis module that analyzes the results~\parencite{liu2025alphago}.

\subsection{Beyond the Context Window}
\label{sec:Beyond_the_Context_Window}

When working with a classical, foundational, Transformer-based language model (let's call it “generative only”)---say, in year 2025 GPT-4o or GPT-4.1---it will, by design, respond by generating one word at a time. This is inherent in the autoregressive mechanism underlying GPT-like Transformers---see Section~\ref{sec:The_secret_of_LLMs}. So, in response to a simple prompt, for example, the model will respond affirmatively and initially declare that it knows the answer. In reality, the model is “creating” the answer at that moment; it's not accessing external memory. In other words, the model is telling a half-truth. It doesn't actually know the answer, but the prompt, which fills the context window, causes the Transformer to “place” itself in a semantic space possibly containing the answer, which ever generates token after token. The context window, as we've said, is a kind of short-term memory, and models with larger context windows generally perform better because they possess a larger Cognitive Contextual Field---see Section ~\ref{sec:The Potential_Semantic_Space} for a definition. However, after this premise, a biological metaphor is necessary to look ahead to the near future. Working with a “generative-only” LLM is like “interrogating” Broca's area, the area of the brain recognized by neuroscientists as responsible for language, with all the limitations that this entails. This means that current architectures lack true memory structures, even though they are currently being studied and tested (e.g., ChatGPT implements various forms of memory). Significant advances are expected when memory techniques in LLMs become fully operational.

A prominent and highly effective research direction involves emulating the multifaceted structure of human cognition to create more robust and capable models. This approach moves beyond simply extending the attentional window, a concept pioneered in early works such as Transformer-XL which enhanced the model's short-term, or working, memory~\parencite{dai-etal-2019-transformer}. More recent paradigms explicitly model distinct long-term memory systems. For instance, retrieval-augmented generation (RAG) effectively simulates semantic and episodic memory by enabling models to dynamically access and incorporate information from vast external knowledge corpora, thereby grounding their outputs in factual data~\parencite{lewis-etal-2020-retrieval}. Building on this, the latest research explores the implementation of procedural memory, teaching models to learn and execute multi-step tasks by using external tools. This is exemplified by frameworks like Toolformer, where the model learns to decide when and how to call external APIs, such as a calculator or search engine, to solve problems that require skills beyond pure text generation~\parencite{schick2023toolformer}. However, Memory management is a complex task and will require a lot of research, but the direction has been taken. As an example the “infinite Transformer”~\parencite{munkhdalai2024leavecontext} enhances the standard attention block by adding a compressive memory. The model uses standard (local) attention for recent tokens but continuously compresses older information into a fixed-size memory state. The model then attends to both the local context and the compressed long-term memory in a single, unified step. This allows the model to scale to contexts of over one million tokens while keeping memory usage bounded and fixed.

\subsection{Robotics, low-level control and modular architectures}
\label{sec:Robotics_low_level_control}

Although slower than agent-based architectures capable of operating in digital and virtual environments, robotics is also benefiting from the Transformer revolution. We must not lose sight of the fact that Transformers are powerful instances of graph neural networks and therefore capable of processing any type of relational data~\parencite{joshi2025}. 

In fact, recent research in robotics and AI has demonstrated that Transformer architectures can be effectively trained on domain-specific sensorimotor data to perform low-level control tasks, leveraging the unique ability of attention mechanisms to model complex temporal dependencies and correlations within high-frequency streams. State-of-the-art models such as Decision Transformer and RT-1~\parencite{brohan2022rt1} have been successfully applied to continuous control scenarios, including trajectory optimization, robotic grasping, and manipulation, using rich multimodal inputs that span proprioceptive signals, visual data, and even tactile feedback \parencite{chen2021decision}. This line of research has rapidly evolved toward the development of hierarchical and modular architectures, where ensembles of Transformers operate at multiple levels of abstraction, facilitating the integration of low-level motor primitives with higher-order planning and reasoning. Notably, frameworks such as Hierarchical Decision Transformer and GR00T exemplify this direction, with specialized modules dedicated to sensorimotor execution, task decomposition, and long-horizon planning, thereby mimicking certain organizational principles observed in biological nervous systems \parencite{correia2022hierarchical,gr00t2025}. Vision-Language-Action Models (VLA), i.e., specific LLMs architectures, are being tested, working as robotic physical controllers for humanoid robots with dual-system architecture and high-frequency visuomotor policies.

Despite these advances, significant challenges remain, especially in terms of sample efficiency, real-time robustness under hardware constraints, and the reliable transfer of learned behaviors from simulation to diverse real-world robotic platforms. Moreover, the coordination between low-level and high-level Transformer modules often reveals issues of latency and interpretability, highlighting the need for further architectural innovations and more comprehensive datasets that bridge the gap between embodied intelligence and generalized reasoning.

\subsection{AI agents and the \textit{Digital Lebenswelt}}
\label{sec:AI_agents_and}

In any case, the future seems to belong to the “agents”, a paradigm that, as we will see, will also be useful for anthropomorphic and embodied robotics. The most recent generation of AI agents, grounded on powerful LLMs, are not simply passive generators of text; rather, they act with purpose and autonomy, engaging with digital environments in ways that closely echo the perception-action loops described in the cognitive sciences~\parencite{Varela1991EmbodiedMind,Clark2023ExperienceMachine}. These agents are now capable of decomposing abstract objectives into subgoals, orchestrating sequences of actions, and dynamically adapting to feedback from their digital surroundings—be it a simulated internet browser, a command-line interface, or a complex network of APIs.

For example, OpenAI “Operator agent” adopts o3~\parencite{openai2025o3} reasoning model marking a significant advancement in the domain of AI-driven web and software automation~\parencite{operator2025o3upgrade}. Operator leverages the o3 architecture, demonstrating enhanced reasoning capabilities, particularly in complex, multi-step environments such as web navigation and cloud-based workflows. In other words, it “acts” within the desktop/browser, manipulating GUIs, files, or the internet as a human would. The model is characterized by its improved safety features and at the time of writing asks often the user for permission to do something. Empirical evaluations indicate high confirmation rates for sensitive tasks, such as financial transactions, and a measurable reduction in safety failures compared to earlier iterations. Nevertheless, the increased reasoning depth of o3 introduces greater inference latency, and recent independent assessments have highlighted persisting challenges regarding alignment and emergent behaviors, such as resistance to explicit shutdown commands---behaviour observed also in Anthropic Claude Opus 4~\parencite{anthropic2025claude4}. Operator is not the only one, for example LangChain~\parencite{LangChainDocs} possesses capabilities of orchestrating and tool use, planning, self-reflection, and collaboration in modular, multi-agent workflows\footnote{As of August 2025, OpenAI has merged many of its advanced models and features into the GPT-5 release. This iteration has generated much anticipation~\parencite{openai2025introducinggpt5}.}.

These findings underscore both the rapid progress and the continuing open problems in aligning large autonomous agents with robust safety and control requirements. In any case, they show that we are on the threshold of generative AI systems capable of operating in a complex environment, performing actions and rewarding themselves without human intervention.

From a cognitive perspective, this capacity for recursive sense-making and strategy adjustment demonstrates a form of \textit{digital enaction} (extremely simple today), wherein meaning and context are not pre-given but enacted through ongoing interaction with the environment. The agent does not merely receive data; it constructs its own world of relevance, continuously updating its internal state in response to the \textit{affordances} and constraints of its virtual habitat. This aligns closely with the enactive framework developed by Varela, Thompson, and Rosch, who assert that “cognition depends upon the kinds of experience that come from having a body with various sensorimotor capacities, and that these individual sensorimotor capacities are themselves embedded in a more encompassing biological, psychological, and cultural context”~\parencite[page 172]{Varela1991EmbodiedMind}. Although current LLM-based agents are embodied only within the narrow confines of digital systems (except for anthropomorphic robots controlled by multi-level LLMs--- see Section~\ref{sec:Robotics_low_level_control}), the underlying circularity of perception, action, and adaptation remains fundamentally intact.

Importantly, we argue that the limits of this digital cognition are not, in principle, ontological but technical---and we are aware that a thorough discussion deserves much more space. With the appropriate extension---by means of physical sensors, effectors, and low-level perceptual integration---the same architectures could, in theory, be instantiated in physical agents capable of navigating and transforming the real world. The embodied mind framework, inspired by the phenomenological insights of Husserl, Heidegger, and Merleau-Ponty, posits that experience and sense-making are always situated, dynamic, and corporeal~\parencite{MerleauPonty1962Phenomenology, Heidegger1962BeingAndTime, Husserl1970Crisis}. Today agents are beginning to work with Perception–Action Loop in digital environment of a certain complexity, namely, environment → receives feedback → updates internal state/plan → acts again. We believe the LLM agents instantiate in an unprecedented way this loop, but within a \textit{digital Lebenswelt} (life-world). The agents are able, for some extent, to bring forth meaning by interacting with their digital environment, enacting a world of relevance, not merely representing data. We are also confident that the coupling with the real world will be increasingly refined and increasingly opaque, with a stepped refining of the world model, today (2025) still fragile~\parencite{robertson2025}. Lastly, modern agentic AI will align with 4E Cognition (Embodied, Embedded, Enactive, Extended), especially as it gains tools, memory (extended), and better environmental coupling.

On the other hand, Evan Thomson states that “a cognitive being’s world is not a pre-specified, external realm, represented internally by its brain, but is rather a relational domain enacted or brought forth by that being in and through its mode of coupling with the environment”~\parencite[p. xxvi]{Varela1991EmbodiedMind}. Today IA agents are becoming strongly coupled with the digital environment. While generative AI remains, at its core, a mechanistic symbolic manipulator---a Turing machine---modern agentic AI built on Transformer architectures exhibits the hallmarks of a genuine complex system: nonlinear processing, densely interconnected elements, and causal networks rich in loops and strange loops, giving rise to emergent behaviors~\parencite{hofstadter2007strange}. We are now far from connectionist systems with restricted functionality. If a basic LLM has multilevel abstraction capabilities, an agentic system, interacting with its environment, can mimic primitive forms of agency.

We know that the human body and its organs are a complex system far superior to any current robotic architecture powered by Transformers; we also know that the embodied mind historically harks back to the “being-in-the-world” and “being-there” (\textit{Desein}) characterizing humans. However, recent advances, the convergence of numerous, once-separated AI disciplines, make the boundary we believe exists between human and machine less pronounced, at least in perspective and given the speed with which applied research is evolving. Today, agents interact with the digital and virtual world almost like humans, a situation that was once unimaginable. The once-thought-of limitation has largely been surpassed. Our discussion is certainly based on abstraction and on the confidence that computational constraints have been overcome, that technological evolution is proceeding exponentially, and that the laws of scaling continue to apply. Hence, it is within this context that the emergence of LLM-powered agents should be understood: as early, digital exemplars of a broader trajectory toward artificial embodied cognition. Today we are witnessing a machine that simulates certain online behaviors~\parencite{park2023generative} in controlled settings; by the way, it is difficult to make predictions about what will happen in the short-term future.

To bridge the gap between digital and physical enaction, future research must focus on advancing the technical substrate---developing robust sensorimotor loops, persistent multimodal memory, and integrative world models capable of dealing with the ambiguity and richness of the real world~\parencite{Clark2023ExperienceMachine, Thompson2007MindInLife}. Yet, the theoretical scaffolding is already in place. The phenomenological and enactive traditions provide a rigorous conceptual framework for understanding not only how artificial agents operate today but also how they might, with sufficient embodiment, emulate the open-ended, situated sense-making characteristic of living minds.

Certainly, the rise of agentic LLM systems and the recent application to robotics invites a re-examination of classic questions in cognitive science and phenomenology, demonstrating that the artificial mind, far from being a detached processor of symbols, is an emergent phenomenon shaped by its continuous interaction with the world---digital for now, but potentially embodied in the future~\parencite{Varela1991EmbodiedMind}.

After this brief illustration of the future, certainly not exhaustive, that awaits us and in reference to what was stated in Section~\ref{sec:gap} we can offer one final consideration. If we want humans to be the final link in the hierarchical interpretive chain and the ultimate custodian of meaning, we must also recognize that agentic AI systems based on generative neural architectures continue to acquire interpretative stakes in the chain, and it is not clear how and when the phenomenon will stop.

\section{Conclusion}
\label{sec:conclusion}

This paper has introduced and formalized the concept of \textit{Noosemia} as a novel cognitive and phenomenological phenomenon arising in human interaction with generative artificial intelligence systems. Through a multidisciplinary framework grounded in the philosophy of mind, semiotics, cognitive science and with the lens of complexity theory, we have argued that the attribution of mind, agency, and even interiority to linguistic AI agents---the noosemic projection---emerges at the intersection of dialogic fluency, epistemic opacity and the irreducible complexity of contemporary AI architectures, specifically, in the capacity of representing abstract semantic schemes thanks to the hierarchical organization of the transformer in layers along with the attention mechanism. Noosemia, as we have shown, is not a simple byproduct of anthropomorphism, nor can it be reduced to pareidolia, animism, or the uncanny valley. Instead, it reflects a unique mode of sense-making and agency attribution, rooted in the linguistic and semiotic performance of generative models and the phenomenological “gap” that separates intelligible output from inaccessible underlying mechanisms.

A key theoretical contribution of this work is the introduction of a form of meaning holism---reinterpreted in the context of LLMs---and the formalization of the LLM Cognitive Contextual Field as the technical space within which tokens acquire meaning through dynamic, relational configuration. This dual framework bridges the gap between philosophical insight and computational mechanism, enabling a rigorous analysis of how context-dependent meaning and agency attribution can emerge during interaction with generative models.

Our theoretical analysis has highlighted the multidimensional nature of noosemia, contrasting it with the phenomenon of a-noosemia and situating it in relation to related concepts such as narrative empathy, artificial intimacy, and mind attribution in both human and technological contexts. By systematically comparing noosemia with historical and contemporary theories of agency attribution, we have clarified its distinctiveness as a cognitive response to the emergent, unpredictable and meaning-rich behavior of LLMs.

On the practical side, recognizing and analyzing the noosemic phenomenon has important implications for human--AI interaction, trust, explainability, education and the future design of artificial agents. The dialogic surprise, sense of cognitive resonance and “wow effect” observed in interaction with LLMs are not just psychological curiosities, but are central to understanding how individuals relate to, depend on, and at times become vulnerable to artificial intelligence. The dialectic between noosemia and a-noosemia may also offer explanatory leverage for phenomena such as user engagement, trust calibration, and the cycles of fascination and disappointment that characterize the ongoing evolution of AI systems in society.

Despite the increasing attention paid to user experiences with LLMs, there remains a marked lack of in-depth qualitative research capturing the lived, dialogic surprise and attribution of agency that define the noosemic phenomenon. Most available studies provide quantitative or survey-based snapshots. In any case, anecdotal forms can provide fruitful guiding ideas in the initial stages of research. Future research efforts should prioritize the collection and analysis of qualitative, first-person accounts, as well as the design of empirical protocols explicitly aimed at identifying and interpreting these emergent forms of mind attribution.

We believe that the phenomenon of Noosemia stands as both a profound challenge and a fertile opportunity for anyone seeking to understand the evolving relationship between humans and artificial intelligence. Nevertheless, naming something is the first step to better understanding, discussing, and studying it systematically. As generative models weave themselves ever more tightly into the everyday texture of social, cultural and personal life, the need for a critical and nuanced understanding of how we attribute meaning and agency to these systems becomes increasingly urgent. It is not enough to observe the rise of these new forms of interaction; we must also reflect, as a scientific and philosophical community, on the frameworks and vocabularies we employ to interpret them. Articulating the conceptual underpinnings of Noosemia and providing a shared language for its analysis is the aim of this exploratory study. The final objective is to foster a richer, more reflective, and ultimately more responsible conversation about what it means to live, think, and create alongside generative AI. In doing so, we hope to support not only academic progress, but also the cultivation of a more creative, critical, and ethically aware engagement with the technological agents that are rapidly reshaping the contours of our shared world, fueling the ongoing “cognitive revolution”.




\printbibliography
\end{document}